\documentclass{article}

\usepackage{PRIMEarxiv}

\usepackage[utf8]{inputenc} 
\usepackage[T1]{fontenc}    
\usepackage{hyperref}       
\usepackage{url}            
\usepackage{booktabs}       
\usepackage{amsfonts}       
\usepackage{nicefrac}       
\usepackage{microtype}      
\usepackage{lipsum}
\usepackage{fancyhdr}       
\usepackage{graphicx}       
\graphicspath{{media/}}     

\usepackage{bm}
\usepackage{amsmath}
\usepackage{amssymb} 
\usepackage{mathrsfs}
\usepackage{graphicx}
\usepackage{pgfgantt} 
\usepackage[colorinlistoftodos]{todonotes}

\usepackage{caption} 
\usepackage{subcaption}
\usepackage{setspace} 
\usepackage{comment}

\usepackage{float} 

\numberwithin{equation}{section}

\title{Mori-Zwanzig latent space Koopman closure for nonlinear autoencoder
}

\author{
  Priyam Gupta \\
  Department of Aeronautics \\
  Imperial College London \\
  London SW7 2AZ, United Kingdom\\
  \texttt{priyam.gupta21@imperial.ac.uk} \\
   \And
  Peter J. Schmid \\
  Department of Mechanical Engineering \\
  KAUST \\
  23955 Thuwal, Saudi Arabia\\
    \And
  Denis Sipp \\
  DAAA \\
  Onera \\
  92190 Meudon, France\\
    \And
  Taraneh Sayadi\\
  Institut Jean le Rond d'Alembert, Sorbonne University\\
  75005, Paris, France\\
  Institute for Combustion Technology, Aachen University\\ 52062 Aachen, Germany
    \And
  Georgios Rigas \\
  Department of Aeronautics \\
  Imperial College London \\
  London SW7 2AZ, United Kingdom\\
}

\begin{document}
\maketitle

\begin{abstract}

The Koopman operator presents an attractive approach to achieve global linearization of nonlinear systems, making it a valuable method for simplifying the understanding of complex dynamics. While data-driven methodologies have exhibited promise in approximating finite Koopman operators, they grapple with various challenges, such as the judicious selection of observables, dimensionality reduction, and the ability to predict complex system behaviours accurately. This study presents a novel approach termed Mori-Zwanzig autoencoder (MZ-AE) to robustly approximate the Koopman operator in low-dimensional spaces. The proposed method leverages a nonlinear autoencoder to extract key observables for approximating a finite invariant Koopman subspace and integrates a non-Markovian correction mechanism using the Mori-Zwanzig formalism. Consequently, this approach yields an approximate closure of the dynamics within the latent
manifold of the nonlinear autoencoder, thereby enhancing the accuracy and stability of the Koopman operator approximation.  Demonstrations showcase the technique's improved predictive capability for flow around a cylinder. It also provides a low dimensional approximation for Kuramoto-Sivashinsky (KS) with promising short-term predictability and robust long-term statistical performance. By bridging the gap between data-driven techniques and the mathematical foundations of Koopman theory, MZ-AE offers a promising avenue for improved understanding and prediction of complex nonlinear dynamics.

\end{abstract}


\section{Introduction}\label{sec:intro}

Nonlinear systems are ubiquitous, spanning from unsteady fluid flows and the evolution of epidemics to intricate neural interactions in the brain. These systems frequently exhibit a level of dimensionality that is exceedingly high for any practical computational analysis. As a pragmatic solution, reduced order models (ROM) are sought which provide tractable and accurate dynamics for a small set of quantities of interest. Since the intricate dynamics exhibited by these systems arise from the nonlinear spatio-temporal interactions between multiple scales, it often makes it difficult to find their analytical solutions. The Koopman operator \cite{koopman1932dynamical, mezic2005spectral} offers an alternative view through the lens of ``dynamics of observables'' which facilitates a global linearisation for these inherently nonlinear systems. This linear characteristic holds significant appeal across a spectrum of applications, including nonlinear system identification \cite{klus2020data} and nonlinear control \cite{kaiser2021data}. However, the Koopman operator's definition within an infinite-dimensional Hilbert space contradicts the fundamental goal of constructing a ROM. Therefore, we seek a finite low-dimensional representation which entails identifying a specific set of observables (resolved observables) that span a finite invariant Koopman subspace. This task is often challenging and requires approximation \cite{brunton2016koopman}. When approximating, one must anticipate the accumulation of errors in the generated state-space trajectories over time. The Mori-Zwanzig formalism \cite{mori1965transport,zwanzig1973nonlinear} enables a closed-form representation of the infinite-dimensional Koopman operator by providing a non-Markovian correction to its finite approximation. This effectively closes the gap between the practical implementation of the Koopman operator and its idealised form.

 Dynamic mode decomposition (DMD) has shown a remarkable performance for approximating the Koopman operator for fluid flows~\cite{page2019koopman,sayadi2014reduced, korda2020data}, or many other dynamical systems, for example, epidemic evolution~\cite{proctor2015discovering}. Nonetheless, it struggles with the issue of choosing a proper set of observables. This is mainly because deriving the closed-form solution in the Koopman theory involves carefully resolving the key observables enriched in the nonlinear information of the system and embed them into a linear dynamics. One approach to identify these observables is to search for them within a pre-defined dictionary (referred to as a feature map) of linear and nonlinear functions of the state variables, as proposed by Williams~\emph{et al.}~\cite{williams2015data}. This extended DMD (EDMD) method is equivalent to using a high-order Taylor series expansion around equilibrium points as compared to merely a linear expansion by standard DMD~\cite{schmid2022dynamic}. However, this approach relies on \textit{a priori} knowledge of the behaviour of the dynamical system. Another caveat of the same approach is the limited representational capacity of the dictionary which can lead to overfitting due to insufficient data.

An alternative technique for approximating the Koopman operator is to exploit the universal approximation capability of the neural networks to learn the observables from data. This method is generally referred to as the Deep Koopman approach \cite{otto2019linearly, lusch2018deep, takeishi2017learning, yeung2019learning} where the state variables are
passed through a nonlinear autoencoder to produce a small set of observables enriched with the nonlinearities of the dynamical system. To ensure the observables lie in the linearly invariant subspace, an approximate Koopman operator is obtained through linear regression in time over these observables. The motivating idea behind this approach is a two-step identification where (i) the autoencoder learns energetically dominant modes, and (ii) the approximate Koopman operator learns dynamically important features. Otto \emph{et al.}~\cite{otto2019linearly} used a Linear Recurrent Neural Network framework where the error of the learned Koopman operator is minimized over multiple timesteps.  Lusch \emph{et al.} \cite{lusch2018deep} extended this work to dynamical systems with continuous frequency spectra. They obtained the parametric dependence of the Koopman operator on the continuously varying frequency using an auxiliary network. DMD based approach that involves Moore-Penrose pseudo-inverse for approximating the finite Koopman operator has also been tested with these neural network-based dictionaries \cite{constante2023enhancing}. Pan \emph{et al.}~\cite{pan2020physics} proposed a probabilistic Koopman learning framework based on Bayesian neural networks for continuous dynamical systems while offering a stability constraint on their Koopman parameterization. 

 Data-driven Koopman learning methods are founded on the assumption that a non-trivial {\it{finite}}-dimensional Koopman invariant subspace exists \cite{colbrook2023residual}. Even if this assumption holds true, it has proven to be exceedingly challenging to resolve this finite set of observables that completely closes the dynamics \cite{brunton2016koopman}.  In order to obtain a closed dynamics, we need to account for the effects of the unresolved observables that complete the invariant Koopman subspace. Mori and Zwanzig introduced a general framework for the closed equations of the resolved observables. They demonstrated that the interactions between resolved and unresolved observables manifest themselves as non-Markovian non-local effects on the resolved observables. To accommodate these interactions, it decomposes the dynamics into three parts -- a Markovian term, a non-Markovian or memory term, and a noise term -- which together form a so-called generalized Langevin equation (GLE). In this decomposition, the memory and the noise terms are responsible for the effects of the unresolved observables. While the GLE provides methodically exact evolution equation for resolved observables, it does not provide reduced computational complexity without approximations. This is primarily because deriving the analytical form of the memory kernel which accounts for the non-Markovian effect is an arduous task. However, the GLE provides an excellent starting point to model closure terms in a non-Markovian form.
 
 Mori-Zwanzig formalism has been exploited by several works that propose a higher-order correction to the approximate Koopman operator by accounting for the residual dynamics through the memory term. Lin \emph{et al.}~\cite{lin2021data_2} proposed a data-driven method for this purpose that recursively learns the memory kernels using Mori's linear projection operator. This work was further advanced in~\cite{lin2022regression} through the use of a regression-based projection operator. In this study, neural network-based models were employed to derive linear and non-linear projection operators, rather than to learn a dictionary of observables. Curtis \emph{et al.}~\cite{curtis2021dynamic} used the popular optimal prediction framework\cite{chorin2000optimal} to provide higher-order correction terms for DMD. This was further improved in~\cite{katrutsa2023extension} where $t$-model \cite{chorin2002optimal} was utilized for memory approximation. The primary challenge for these methods is the judicious choice of the observables. The optimal resolved observables are those where dynamics of the system are largely concentrated. Generally, they are selected from a predefined dictionary of functions that has the same shortcomings as mentioned before, such as overfitting and the need for \textit{a priori} knowledge of the system as in  EDMD. This points to a need for autonomous selection of observables from data.
 
Observing these problems, this work proposes an interpretable data-driven ROM termed Mori-Zwanzig autoencoder (MZ-AE), that exploits the Mori-Zwanzig formalism and approximates the invariant Koopman subspace in the latent manifold of a nonlinear autoencoder. A non-Markovian correction is provided to the approximate Koopman operator which guides it back to the true trajectory upon deviation. Through this approach, we tackle the following challenges:

\begin{itemize}

   \item \textbf{Low-dimensional approximation}: For a ROM a tractable number of observables are required that evolve accurately with time on a low-dimensional subspace. An excessive number of observables can lead to spurious eigenvalues which are non-physical and impede interpretability\cite{colbrook2023residual}. A nonlinear autoencoder provides good choice of observables, which, combined with the closure from a non-Markovian correction term enables an aggressively low-dimensional model.
   
  \item \textbf{Choice of observables}: A nonlinear autoencoder provides a data-driven approach to finding the key observables for best approximation of the Koopman operator. This enables the resolved observables to learn the dominant energy observables. Further, this relaxes the requirement of \textit{a priori} knowledge of the system while providing a suitable coordinate transformation into an approximate Koopman invariant subspace.
  
  \item \textbf{Predictability}: It is highly challenging to obtain finite linear approximations for a chaotic system \cite{budivsic2012applied}. These linear models fail to achieve good long-term predictions and require large number of observables to unfold the dynamics further \cite{otto2019linearly} defeating the purpose of an ROM. Through the Mori-Zwanzig formalism, we provide a memory correction term to provide longer predictability.
  
\end{itemize}

\textit{Organization.} \S\ref{sec:BT} presents the relevant background theory, briefly discussing model-order reduction, Koopman formalism, and Mori-Zwanzig decomposition. \S\ref{sec:Methodology} provides the details of the proposed MZ-AE algorithm. \S\ref{sec: Num_Exp} sets up the numerical experiments and discusses the results obtained. Finally, in  \S\ref{sec: Conclusion} we provide conclusions and a discussion on future work.

\section{Background theory}
\label{sec:BT}
This section lays down the mathematical background building up to the framework proposed in this work. We start with Poincare's state space view of the dynamical system in order to set up the operator theoretic method of Koopman. Subsequently, we motivate our approach from the reduced order model perspective since we seek a low-dimensional representation of the infinite Koopman operator. Finally, we discuss the Mori-Zwanzig decomposition of the observable dynamics and provide a rationale for the proposed MZ-AE framework to close the finite Koopman approximation for nonlinear autoencoders.

\subsection{Data-driven model order reduction}

We consider an autonomous dynamical system evolving on a smooth manifold $\mathcal{M} \subseteq \mathbb{R}^N$,

\begin{equation}
    \label{eq:Continuous_Dynamical_System}
    \frac{d\bm{\Phi}(t)}{dt}  = S(\bm{\Phi}(t)), \quad \bm{\Phi}(0) = \bm{\Phi}_0,
\end{equation}
where $ \bm{\Phi}(t) \in \mathcal{M}$ is a $N \times 1$ dimensional \textit{state} vector or \textit{phase-space} vector which characterizes the state of the system at any time $t \in \mathcal{T} = \mathbb{R}$. The evolution operator $S:\mathcal{D}(\mathcal{M})\rightarrow\mathcal{M}$ is a continuously differentiable map, where $\mathcal{D}(\mathcal{M})$ is the domain of $S$. The dynamical system has an associated flow $T:\mathcal{M}\times \mathbb{R} \rightarrow \mathcal{M}$ defined as,
\begin{equation}
    T(\bm{\Phi}_0, t) = \bm{\Phi}_0 + \int_{0}^{t} S(\bm{\Phi}(\tau)) d\tau = \bm{\Phi}(t).
\end{equation}
For practical purposes, we are specifically interested in a discrete autonomous dynamical system sampled at fixed time step $\Delta t$,
\begin{equation}
    \label{eq:Discrete_Dynamical_System}
    \bm{\Phi}_{n+1} = \mathbf{T}_{\Delta}(\bm{\Phi}_{n}), \quad n\in\mathbb{Z},
\end{equation}
where $\bm{\Phi}_{n} = \bm{\Phi}(n\Delta t)$ and $\mathbf{T}_{\Delta}$ is the discrete map for the flow $T$. The evolution operator $S$ is referred to as full order model (FOM) which is typically characterised by non-linearity and the phase-space variables ($\bm{\Phi}$) exhibit high dimensionality. These properties make it prohibitively expensive to obtain the solution for the system and only limited insights can be gained into the inherent system dynamics. To address this, we seek an accurate and amenable ROM for a tractable number of relevant quantities $\mathbf{u}(\bm{\Phi}) \in \mathbb{R}^{r}$ such that $r \ll N$.

A data-driven ROM extracts this low-dimensional representation of the full-order model from the data snapshots. This is done by projecting onto a data-driven expansion basis as in Proper Orthogonal Decomposition (POD)~\cite{lumley1967atmospheric}. In this method the basis functions (or modes) are ranked according to their energy content and the dimension reduction entails decomposing the phase-space variable into first $r$ dominant energy modes,
\begin{equation}
    \label{eq:POD_decomposition}
    {\Phi}(\mathbf{x},t) = \sum_{j=0}^{r} a_j(t){\mathbf{u}}_j(\mathbf{x}) + \mathbf{\epsilon}, 
\end{equation}
where $\mathbf{x}$ represents the spatial coordinates, $\mathbf{u}_j$ is the spatial mode, $a_j$ is the time-dependent amplitude of the spatial modes and $\mathbf{\epsilon}$ is the residual error from unresolved modes. The dynamical system can then be  projected on to these modes using Galerkin projection to obtain a  reduced order model \cite{rowley2017model}. Alternatively, a data-driven model, such as a Recurrent Neural Network (RNN), can be used to learn the projected non-linear dynamics \cite{maulik2020time}. There is also a need to model effects of unresolved modes on the dynamics of resolved modes and provide a closure\cite{wang2020recurrent, menier2023cd}. The key limitation of these methods is that the POD modes tend to intertwine spatial and temporal frequencies which complicates their physical relevance \cite{oberleithner2011three}. As a solution to this limitation, dynamic mode decomposition (DMD) based methods provide a data-driven approach to extract dynamically relevant modes representing spatio-temporal coherent structures by regressing a linear evolution operator motivated by Koopman formulation. 

\subsection{Koopman formalism}

The semigroup of Koopman operators corresponding to the dynamical system (equation~\eqref{eq:Continuous_Dynamical_System}) operates on a separable infinite dimensional Hilbert space, $\mathcal{H} = \mathcal{L}^{2}(\mathcal{M},\mathbb{R})$, of square-integrable observable-functions, $g:\mathcal{M}\rightarrow\mathbb{R}$. This space has an associated inner product $\langle\cdot,\cdot\rangle$. $\mathcal{H}$ is a linear vector space over $\mathbb{R}$, however, the scalar-valued observable-functions (observables) can be linear or non-linear functions of phase-space variable ($\bm{\Phi}$). Starting from the initial condition $\bm{\Phi}_0$, the observable at any time $t \in \mathbb{R}^{+}$ attains the value $g(T(\bm{\Phi}_0, t))$ which we represent by $g(\bm{\Phi}_0,t)$. The Koopman operator $\mathcal{K}:\mathcal{H}\rightarrow\mathcal{H}$ is a bounded linear operator acting on this observable space,

\begin{equation}
    \label{eq: continuous_koopman}
    \mathcal{K}^{t}g(\bm{\Phi}_0) = g \circ T(\bm{\Phi}_0,t).
\end{equation}  
A finite number of observables $g^{(i)}, i \in \{1\cdots p\} $ can be expanded into an infinite-dimensional space spanned by the Koopman eigenfunctions $\Psi_j $, which satisfy  $ \mathcal{K}^t\Psi_j=e^{\lambda_j t}\Psi_j $.
We then have:
\begin{equation}
    \label{eq:eigen-decomposition}
    \mathcal{K}^t g^{(i)}(\bm{\Phi}_0) = \sum_{j=1}^{\infty}v_{i,j}e^{\lambda_j t}\Psi_{j}(\bm{\Phi}_0),
\end{equation}
where the coefficient $v$ is the corresponding Koopman eigenmode, that may be obtained by expressing each observable in the Koopman eigenfunction basis. These eigenfunctions contain the characteristic dynamical features of the system and enable the linearisation of the non-linear systems. 
 
\subsection{Koopman Finite Representation}
In view of obtaining a reduced-order model, an infinite-dimensional space is not tractable. We therefore seek a finite-dimensional subspace of observables $\mathcal{F}=\mbox{Span}\{g^{(i)}:\mathcal{M}\rightarrow\mathbb{R}\}_{i=1}^{r}$ (limited to $r \ll N$) and a finite-dimensional representation $\mathbf{K}^t:\mathcal{F}\rightarrow\mathcal{F}$ such that $\mathcal{K}^t\mathbf{g} = \mathbf{K}^t\mathbf{g}$, where $\mathbf{g} = \left[g^{(1)},...g^{(r)}\right]^{T}$. 

A linear diagonalizable finite representation $\mathbf{K}^t=\mathbf{V}e^{\Lambda t}\mathbf{V}^{-1}$ generates a finite number of observables  $\mathbf{h}=\mathbf{V}^{-1} \mathbf{g}$, which are eigenfunctions of $\mathcal{K}^t$, since $\mathcal{K}^t\mathbf{h}=\mathbf{h} \circ T =\mathbf{V}^{-1} \mathbf{g}\circ T=\mathbf{V}^{-1} \mathbf{K}^t\mathbf{g}= e^{\Lambda t} \mathbf{h}$. The objective boils down to ascertaining the observables and obtaining the matrix $ \mathbf{K}^t $ that spans a finite closed invariant subspace, such that $\mathbf{g} \circ T = \mathbf{K}^t\mathbf{g}$. Reciprocally, a finite collection of Koopman eigenfunctions generates a finite representation $ \mathbf{K}^t $. 

In practice, it is challenging to find such a finite closed invariant subspace and we are limited to an approximate finite representation $\tilde{\mathbf{K}}:\hat{\mathcal{H}}\rightarrow\hat{\mathcal{H}}$ acting on the resolved subspace $\hat{\mathcal{H}} = \mbox{Span} \{\hat{g}^{(i)}:\mathcal{M}\rightarrow\mathbb{R}\}_{i=1}^r$. The evolution in this approximation of the invariant subspace ($\hat{\mathcal{H}} \approx \mathcal{F}$) results in the accumulation of errors in the trajectory over time due to loss of dynamical information such that,
\begin{equation}
    \hat{\mathbf{g}}\circ T = \tilde{\mathbf{K}}\hat{\mathbf{g}} + \mathbf{r},
\end{equation}
where $\hat{\mathbf{g}} = \left[ \hat{g}^{(1)}, ...\hat{g}^{(r)}\right]^{T}$ and $\mathbf{r}\in\tilde{\mathcal{H}}$ is the residual residing in unresolved subspace $\tilde{\mathcal{H}}$. This leads to a closure problem \cite{brunton2016koopman}. The Mori-Zwanzig formalism provides a framework to obtain a closed-form equation for this approximate finite representation $(\tilde{\mathbf{K}})$ of the invariant Koopman subspace. This is achieved through an orthogonal projection onto the finite resolved subspace ($\hat{\mathcal{H}}$), while accounting for the effects of the remaining unresolved subspace $( \mathcal{\tilde{H}} )$, where $\mathcal{H} = \hat{\mathcal{H}} \oplus \tilde{\mathcal{H}}$.      

\subsection{Mori-Zwanzig decomposition}
\label{sec: Mori-Zwanzig decomposition}

Let us begin with the continuous formulation and subsequently, adopt its discretized counterpart. We consider an infinitesimal generator (Lie generator) of the Koopman semigroup, called the Liouville operator. It is a linear operator $\mathcal{L}:\mathcal{D}(\mathcal{L})\rightarrow\mathcal{H}$, where $\mathcal{D}(\mathcal{L})\subseteq\mathcal{H}$ which acts on an observable $g$ as,
\begin{equation}
\label{eq:Koopman_semigroup}
   \frac{\partial}{\partial t}g(\bm{\Phi}_0, t) =
 \mathcal{L}g(\bm{\Phi}_0, t), \;\;\;\; g(\bm{\Phi}_0,t) = e^{t\mathcal{L}}g(\bm{\Phi}_0).
\end{equation}
The Liouville operator takes the form
$\mathcal{L}g = S(\bm{\Phi}_{0}) \cdot \left. \nabla_{\bm{\Phi}}g\right|_{\bm{\Phi}_0}$
  which can be obtained by taking the chain time derivative of the observables $g(\bm{\Phi}_{0},t)$ at $t=0$. Note that the Liouville operator depends on the initial state $ \bm{\Phi}_{0}$, and the same operator governs the time-derivative of the observables along the whole trajectory as long as no singularities are encountered.

Hence, we have an explicit expression of the Koopman operator:
\begin{equation}
  \mathcal{K}^{t}g(\bm{\Phi}_0)= e^{t\mathcal{L}}g(\bm{\Phi}_0),
\end{equation}

As discussed before, the Hilbert space is decomposed into a finite resolved subspace $\hat{\mathcal{H}} = \mbox{Span} \{\hat{g}^{(i)}:\mathcal{M}\rightarrow\mathbb{R}
\}_{i=1}^r$ spanned by resolved observables $\hat{\mathbf{g}}$ and an infinite unresolved subspace $\tilde{\mathcal{H}}= \mbox{Span} \{\tilde{g}^{(i)}:\mathcal{M}\rightarrow\mathbb{R}
\}_{i=1}^\infty$ spanned by unresolved observables $\tilde{\mathbf{g}}$ such that $\mathcal{H} = \hat{\mathcal{H}}\oplus\tilde{\mathcal{H}}$. The resolved subspace forms the image of an orthogonal projection map $\mathcal{P}:\mathcal{H}\rightarrow\mathcal{\hat{H}}$. The requirement for this projection is its idempotent nature, characterized by $\mathcal{P}^{2} = \mathcal{P}$. Naturally, there exists a complementary projection $\mathcal{Q} = \mathcal{I} - \mathcal{P}$ whose image $\tilde{\mathcal{H}}$ is the kernel of $\mathcal{P}$. Here, $\mathcal{I}$ represents the identity map. This means that the resolved and unresolved observables should satisfy $\langle \hat{g}^{(i)},\tilde{g}^{(j)}\rangle = 0$, where $\langle\cdot,\cdot\rangle$ is the inner product defined in the Hilbert space. However, we do not require $\langle \hat{g}^{(i)},\hat{g}^{(j)}\rangle=0$, i.e. the orthogonality amongst the resolved observables is not necessary. The nature of the projection operator determines the linear or non-linear nature of the final result of Mori-Zwanzig decomposition which is the GLE. We refer the readers to~\cite{dominy2017duality} for a detailed mathematical treatment of the projection operators in Mori-Zwanzig formulation. 

We are only interested in the evolution of the resolved observables over time i.e. $\hat{\mathbf{g}}(\bm{\Phi}_0,t)$, whose dynamics is described by equation (\ref{eq:Koopman_semigroup}). While these observables start in the resolved subspace, $\hat{\mathbf{g}}(\bm{\Phi}_{0}) \in \hat{\mathcal{H}}$, the action of $e^{t\mathcal{L}}$ provides a transformation in the Hilbert space which pushes them out of $\hat{\mathcal{H}}$ such that $\hat{\mathbf{g}}(\bm{\Phi}_{0},t) \notin \hat{\mathcal{H}}$ as soon as $t>0$. These observables absorb some contribution from the unresolved observables over time. Therefore, the time rate of change of the observables at any time $t \in \mathbb{R}^{+}$ can be decomposed into a resolved and an unresolved part,

\begin{equation}
  \label{eq:liouville_decomposition}
  \frac{\partial }{\partial t} e^{t\mathcal{L}} \hat{\mathbf{g}}(\bm{\Phi}_0) =
  e^{t\mathcal{L}}(\mathcal{P} + \mathcal{Q})\mathcal{L}\hat{\mathbf{g}}(\bm{\Phi}_0) =
  e^{t\mathcal{L}}\mathcal{P}\mathcal{L}\hat{\mathbf{g}}(\bm{\Phi}_0) +
  e^{t\mathcal{L}}\mathcal{Q}\mathcal{L}\hat{\mathbf{g}}(\bm{\Phi}_0).
\end{equation}
The second term on the right-hand side of
equation~(\ref{eq:liouville_decomposition}) needs to be processed to extract the interaction between resolved and unresolved dynamics. This is achieved by applying Dyson's identity, 
%
\begin{equation}
  \label{eq:Applied_Dyson_Identity}
  e^{t\mathcal{L}} = \int_{0}^{t}
  e^{(t-s)\mathcal{L}}\mathcal{P}\mathcal{L}e^{s\mathcal{Q}\mathcal{L}}
  ds + e^{t\mathcal{Q}\mathcal{L}},
\end{equation}
on the unresolved dynamics
$\mathcal{Q}\mathcal{L}\hat{\mathbf{g}}(\bm{\Phi}_0)$ within
equation~(\ref{eq:liouville_decomposition}). Then, using equation~(\ref{eq:Koopman_semigroup}), we obtain the governing
equation which represents the exact evolution of the observable $\hat{\mathbf{g}}(\bm{\Phi}_0, t)$,
\begin{equation}
  \label{eq:Mori_Zwanzig}
  \frac{\partial}{\partial t}\hat{\mathbf{g}}(\bm{\Phi}_0, t) =
  \underbrace{\mathcal{P}\mathcal{L}\hat{\mathbf{g}}(\bm{\Phi}_0,t)}_{\text{Markov}} + \underbrace{\int_{0}^{t}
\mathcal{P}\mathcal{L}e^{s\mathcal{Q}\mathcal{L}}
  \mathcal{Q}\mathcal{L}\hat{\mathbf{g}}(\bm{\Phi}_0, t-s) ds}_{\text{Memory}} +
  \underbrace{e^{t\mathcal{Q}\mathcal{L}}\mathcal{Q}\mathcal{L}\hat{\mathbf{g}}(\bm{\Phi}_0)}_{\text{Noise}}.
\end{equation}

This expression is a formal rewriting of equation \eqref{eq:Koopman_semigroup} for resolved observables. It decomposes the dynamics in infinite-dimensional Hilbert space into a resolved part (Markov term), an unresolved part (noise term), and the interaction between them (memory term). The first term which accounts for the resolved dynamics relies solely on their instantaneous values, hence it is referred to as the Markov term ($\mathbf{M} = \mathcal{PL}$). The third term  $\mathbf{F}(t) =
\mathcal{Q}\mathcal{L} e^{t\mathcal{Q}\mathcal{L}}\hat{\mathbf{g}}(\bm{\Phi}_0)$, incorporates the dynamics of the observables in the unresolved subspace. In literature, it is referred to as the noise term due to its resemblance to the Langevin noise in Langevin equations.  It is apparent that the memory term corresponds to a convolution in time of observables in the past $ \hat{\mathbf{g}}(t-s) $ through a kernel $ \bm{\Omega}(s) = \mathcal{P}\mathcal{L}e^{s\mathcal{Q}\mathcal{L}}
  \mathcal{Q}\mathcal{L}$, in which we recognize part of the noise term $ \mathbf{F}(s)$. This dependence of memory term on noise term is referred to as generalized fluctuation-dissipation (GFD) relation. Consequently, the memory term accounts for the interaction of the resolved and unresolved observables. Equation~\eqref{eq:Mori_Zwanzig} also shows that, given a linear projection operator, the convolution term is, in principle, linear with respect to the past measurements $ \hat{\mathbf{g}}(\bm{\Phi}_0,t-s) $ and that the noise term is orthogonal to $ \mathcal{P}$ (since $\mathcal{P}\mathcal{Q}=0 $). Using the above notations, we rewrite equation~(\ref{eq:Mori_Zwanzig}) in a simplified form as
\begin{equation}
  \label{eq:simplified_Mori_Zwanzig}
  \frac{\partial}{\partial t}\hat{\mathbf{g}}(\bm{\Phi}_0, t) =
  \mathbf{M}\hat{\mathbf{g}}(\bm{\Phi}_0, t) + \int_{0}^{t}
  \bm{\Omega}(s)\hat{\mathbf{g}}(t-s) ds + \mathbf{F}(t).
\end{equation}

This equation is the final result of the Mori-Zwanzig formalism and is often referred to as the generalized Langevin equation. It gives the exact linear evolution of the resolved observables, and consequently the non-linear evolution of the phase-space variables. However, in practice, this equation is not fully closed since, the dynamics of unresolved observables, i.e., the noise term is unknown. This term is often modelled as noise or is neglected. 
\subsection{Discrete GLE}
\label{ref: Discrete GLE}
Since the data for the dynamical systems is often available in discrete format, we are interested in the discrete counterpart of equation~\eqref{eq:simplified_Mori_Zwanzig}. For the discrete system (\ref{eq:Discrete_Dynamical_System}), the Koopman operator takes the form 
\begin{equation}
    \label{eq: discrete_koopman}
    \mathcal{K}_{\Delta}\mathbf{g}_n = \mathbf{g}(\mathbf{T}_{\Delta}(\bm{\Phi}_n)) = \mathbf{g}_{n+1}, 
\end{equation}

where $\mathbf{g}_n = \mathbf{g}
(\bm{\Phi}_n)$. For an approximate finite-dimensional representation of the Koopman operator, we follow the derivations from \cite{darve2009computing, lin2021data}, and obtain the discrete GLE for resolved observables ($\hat{\mathbf{g}}$) by induction,
\begin{equation}
   \label{eq:Discrete_Projected_Mori_Zwanzig}
\mathbf{\hat{g}}_{n+1}=\mathbf{M}_{\Delta}\mathbf{\hat{g}}_{n}+\sum_{k=1}^{n} \bm{\Omega}_{\Delta(k)} \mathbf{\hat{g}}_{n-k} + \mathbf{F}_{n}.
  \end{equation}
Here, the Markov term $\mathbf{M}_{\Delta} = \mathcal{PK}_{\Delta}$ represents the approximate Koopman operator, obtained as the projection of the infinite-dimensional Koopman operator $\mathcal{K}_{\Delta}:\mathcal{H}\rightarrow \mathcal{H}$ onto the finite resolved subspace, $\mathcal{PK}_{\Delta}:\hat{\mathcal{H}} \rightarrow \hat{\mathcal{H}}$. The memory kernel is given by $\bm{\Omega}_{\Delta(k)}\hat{\mathbf{g}}_{n-k} = \mathcal{P}\mathcal{K}_{\Delta}[\mathcal{QK}_{\Delta}]^{k}\mathcal{K}^{n-k}_{\Delta}\hat{\mathbf{g}}_{0}$, while the noise term is expressed as $\mathbf{F}_{n} = [\mathcal{QK}_{\Delta}]^{n+1}\hat{\mathbf{g}}_{0}$. Further, the GFD which, relates the memory term and the noise term can be written in discrete form as,

\begin{equation}
    \label{eq: in text GFD}
    \bm{\Omega}_{\Delta(n)} \mathbf{\hat{g}}_{0} = \mathcal{P}\mathcal{K}_{\Delta}\mathbf{F}_{n-1}, \forall n \in \mathbb{N}.
\end{equation}

When $\mathbf{\hat{g}}$ resides precisely within a finite invariant Koopman subspace, i.e. $\hat{\mathcal{H}} = \mathcal{F}$ , there are no unresolved dynamics i.e. $(\mathcal{Q}\mathcal{K}_{\Delta}) = \{\mathbf{0}\}$. Consequently, the memory and the noise terms become null. Finding such a finite closed invariant subspace is difficult and might not exist in many cases. Generally, we have an approximate finite representation $(\mathbf{\tilde{K}}_{\Delta})$, i.e. $\hat{\mathcal{H}}\approx\mathcal{F}$ which leads to a closure problem. The memory and noise terms provide closure by accounting for the effects of the unresolved observables.

\section{Mori-Zwanzig formulation for nonlinear autoencoder (MZ-AE)}
\label{sec:Methodology}
\begin{figure}[t]
      \centering
      \includegraphics[width=\linewidth]{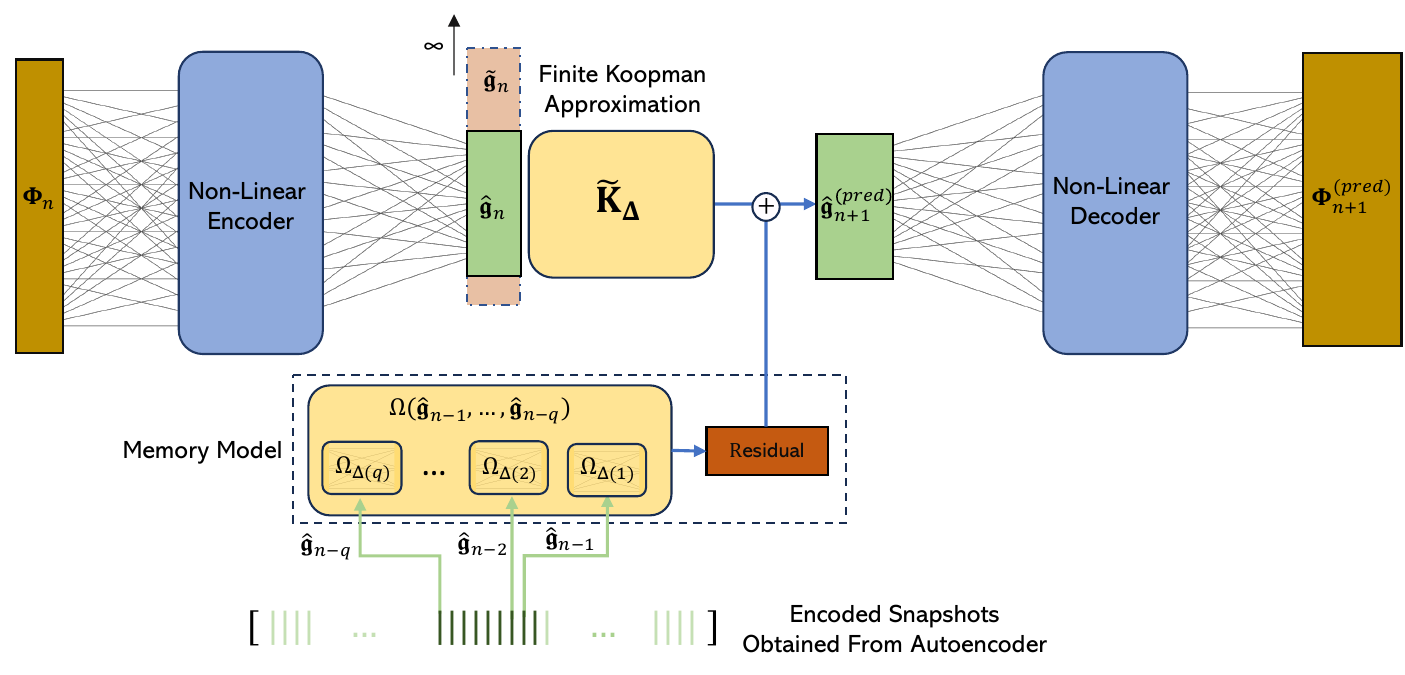}
      \caption{Schematic for the MZ-AE framework.}
      \label{fig:framework}
\end{figure}

 The proposed MZ-AE framework is a data-driven method with threefold objectives, (i) identify the resolved observables $\mathbf{\hat{g}}$ that span an approximate linearly invariant subspace using a non-linear autoencoder, (ii) obtain an approximate finite Koopman representation $(\mathbf{\tilde{K}}_{\Delta})$ that governs the linear evolution of these resolved observables and, (iii) close the dynamics in this approximate invariant subspace by accounting for the effects of the unresolved observables through the memory term (see equation~\eqref{eq:Discrete_Projected_Mori_Zwanzig}). The overall architecture of MZ-AE is shown in Figure \ref{fig:framework}. In this section, we lay down the architecture of the different components used in MZ-AE to achieve the objectives stated above. 

\subsection{Finding resolved subspace using a non-linear autoencoder}

\subsubsection{Non-linear autoencoder}

A nonlinear autoencoder provides a nonlinear coordinate transformation to a low-dimensional latent space. It primarily consists of an encoder and a decoder. The encoder $\mathcal{E}:\mathbb{R}^{N}\rightarrow\mathbb{R}^{r}$ encodes a given input space $\boldsymbol{\Phi} \in \mathbb{R}^{N}$ into the latent space $\mathbf{u} \in \mathbb{R}^{r}$, where $r \in \mathbb{N}$ is the latent space size and $r \ll N$. The decoder $\mathcal{D}:\mathbb{R}^{r}\rightarrow\mathbb{R}^{N}$ remaps this latent space back to the input space by minimizing an appropriate norm (such as $L^2$), giving a reconstruction $\boldsymbol{{\Phi}}^{(pred)}$. The process of encoding and reconstructing is formulated as
\begin{equation}
    \mathbf{u} = \mathcal{E}  \left(\boldsymbol{\Phi};\bm{\theta}_{\mathcal{E}}\right), \quad
    \boldsymbol{\Phi} \approx  \boldsymbol{\Phi}^{(pred)}  = \mathcal{D} \left(\mathbf{u};\bm{\theta}_{\mathcal{D}}\right)
    \label{eq:encoder-decoder eqn}.
\end{equation}

In this study, the encoder and decoder architectures are multilayer perceptron (MLP) networks; however, any other model, such as convolutional neural networks, can be used for the proposed framework. 

\subsubsection{Identifying resolved observables and approximate Koopman operator}

We begin by considering a set of $m$ discrete snapshots of the $N$-dimensional phase-space variables represented by $\bm{\Phi}$. These snapshots can be fully resolved solutions of equation~\eqref{eq:Discrete_Dynamical_System}. The resolved observables  $\hat{\mathbf{g}}:\mathbb{R}^N \rightarrow \mathbb{R}^r$ are the nonlinear functions of these phase-space variables. We find these resolved observables in the latent manifold of a nonlinear autoencoder, $\hat{\mathbf{g}}_n =  \mathcal{E}(\bm{\Phi}_n;\bm{\theta}_{\mathcal{E}})$, sampled at time $\Delta t$. The encoder is responsible for finding the observables that span a linearly invariant subspace. However, as discussed before, we expect to obtain only an approximation of such a space such that,
  \begin{equation}
 \label{eq:simple Koopman}
     \mathbf{\hat{g}}_{n+1} = \tilde{\mathbf{K}}_{\Delta} \mathbf{\hat{g}}_{n} + \mathbf{r}_{n},
 \end{equation}
 where $\mathbf{r}_{n}$ is the residual at the $n^{th}$ timestep. Equation~\eqref{eq:simple Koopman} is equivalent to equation~\eqref{eq:Discrete_Projected_Mori_Zwanzig}, where the approximate Koopman operator $\tilde{\mathbf{K}}_{\Delta}$ represents the Markov term ($\mathbf{M}_{\Delta}$) and corresponds to the projected infinite-dimensional Koopman operator $\left(\tilde{\mathbf{K}}_{\Delta} = \mathcal{PK}_{\Delta}:\hat{\mathcal{H}}\rightarrow\hat{\mathcal{H}}\right)$. The residual $\mathbf{r}_{n}$ comprises both the memory term and the noise term.

To obtain the approximate Koopman operator and the memory kernels, we need a linear projection operator $\mathcal{P}:\mathcal{H} \rightarrow \hat{\mathcal{H}}$. For this purpose, we use linear regression as the projection operator. We begin by considering the data matrix of a batch of $b$ observable vectors at the initial timestep denoted $\mathbf{G}_{0} \in \mathbb{R}^{r \times b}$. These observables correspond to independent resolved observables. Also, consider a data matrix at a future $n^{th}$ timestep denoted $\mathbf{G}_{n} \in \mathbb{R}^{r \times b}$. These observables at the $n^{th}$ timestep are dependent on both initial resolved as well as unresolved observables (recall discussion from \S\ref{sec: Mori-Zwanzig decomposition}, that the observables start from the resolved subspace at $t=0$ and incorporate contributions from the unresolved subspace as they evolve in time). These data matrices are defined as follows,
 \begin{equation}
     \label{eq:data matrices}
     \mathbf{G}_{0} = \begin{bmatrix}
     | & | &  & | \\
     \mathbf{\hat{g}}_{0} & \mathbf{\hat{g}}_{1} & \dots & \mathbf{\hat{g}}_{b-1} \\
     | & | &  & | 
     \end{bmatrix}, \quad
     \mathbf{G}_{n} = \begin{bmatrix}
     | & | &  & | \\
     \mathbf{\hat{g}}_{n} & \mathbf{\hat{g}}_{1+n} & \dots & \mathbf{\hat{g}}_{b-1+n} \\
     | & | &  & | 
     \end{bmatrix}.
 \end{equation}
The linear regression is implemented by minimizing the cost function,
 \begin{subequations}
    \label{eq:linear regression}
    \begin{align} 
     \mathcal{J}(\left\{\mathbf{G}_{n}, \mathbf{G}_{0}\right\} ;\bm{\theta}_{f}) & = \frac{1}{b}||\mathbf{G}_{n} - f(\mathbf{G}_{0}; \bm{\theta}_{f})||^{2}_{F}, \\
     \bm{\theta}^{*} & = \arg\min_{\bm{\theta}_{f}}\mathcal{J}(\left\{\mathbf{G}_{n}, \mathbf{G}_{0}\right\};\bm{\theta}_{f}),
    \end{align}
 \end{subequations}

where $\bm{\theta}_{f} \in \mathbb{R}^{r\times r}$ and $||.||_{F}$ is the Frobenius norm. This linear regression provides an orthogonal linear projection operator $f(\cdot;\bm{\theta}_f)$ that projects $\mathbf{G}_{n}^{T}$ onto the range $\mathcal{R}(\mathbf{G}_{0}^{T})$ such that,
\begin{equation}
    \label{eq: regression projection}
    \mathcal{P}\hat{\mathbf{g}}_{n} (\hat{\mathbf{g}}_{0}) = f(\hat{\mathbf{g}}_{0};\bm{\theta}_f).
\end{equation}

To obtain the approximate Koopman operator as described above and the corresponding resolved observables that span the approximate linear subspace, we simultaneously minimize the following two loss functions,

\begin{itemize}
    \item {\bf{Autoencoder reconstruction loss}}: Minimizing this error
  allows the autoencoder to learn a low-dimensional manifold over which an approximate Koopman subspace can be
  enforced. It is the mean squared error between the
  reconstructed and the actual phase-space variables, stated as 
  \begin{equation}
    \label{eq:Reconstruction_Loss}
    \mathcal{J}_{rec} =
    \frac{1}{b}\sum_{l=0}^{b-1} \left\Vert \boldsymbol{\Phi}_l -
    \mathcal{E}\circ{\mathcal{D}(\boldsymbol{\Phi}_l)}\right\Vert_2^{2}.
  \end{equation}

    \item \textbf{Linear evolution error}: This error enforces a suitable approximate finite Koopman representation $\mathcal{PK}_{\Delta} = \tilde{\mathbf{K}}_{\Delta} (\cdot;\bm{\theta}_{\bm{\tilde{K}}})\in\mathbb{R}^{r\times r}$. To this end, we generate a trajectory over \( \eta \) time steps starting from an initial snapshot \( \hat{\mathbf{g}}_0 \). This trajectory is obtained through the recursive application of the operator 
\( \tilde{\mathbf{K}}_{\Delta} \), resulting in the sequence: $ \left\{ \tilde{\mathbf{K}}_{\Delta}^{j} \hat{\mathbf{g}}_0 \right\}_{j=1}^{\eta}$. The deviation from the true trajectory $ \left\{ \hat{\mathbf{g}}_{j} \right\}_{j=1}^{\eta}$ is quantified by the residual at each timestep $j$ as 
    \begin{equation}
            \label{eq: Koopman Residual}
    \mathbf{R}_{j-1} = \hat{\mathbf{g}}_{j} - \tilde{\mathbf{K}}_{\Delta}^{j} \hat{\mathbf{g}}_0.
    \end{equation}
    
    To evaluate the linear evolution error, we extend this computation to the entire dataset where we use all the snapshots in $\mathbf{G}_{0} \in \mathbb{R}^{r \times b}$ as initial conditions. This generates data matrices of predicted trajectories $\mathbf{G}^{(pred)} \in \mathbb{R}^{r \times b \times \eta}$ and the corresponding true trajectories $\mathbf{G}^{(target)} \in \mathbb{R}^{r \times b \times \eta}$, defined as
    \begin{equation}
     \label{eq:data matrices evolved}
     \mathbf{G}^{(pred)} (:,i,j) = \tilde{\mathbf{K}}_{\Delta}^{j}\hat{\mathbf{g}}_{i}, \quad
     \mathbf{G}^{(target)}(:,i,j) = \hat{\mathbf{g}}_{i+j},
 \end{equation}
    where $i = 0,1,...,b-1$ and $j = 1,2,...,\eta$. 
    The overall linear evolution error $ \mathcal{J}_R $ is measured by the mean squared error of the residuals $ \mathbf{R} $ across all trajectories as
    \begin{subequations}
        \begin{align}
            \mathbf{R}(:,i,j-1) & = \mathbf{G}^{(target)}(:,i,j) - \mathbf{G}^{(pred)}(:,i,j), \label{eq: Linear Evolution residual data} \\
            \mathcal{J}_R (\bm{\theta}_{\tilde{\mathbf{K}}}) & = \frac{1}{b\eta}||\mathbf{R}||_{F}^{2}. \label{eq: Koopman Evolution Error}
        \end{align}
    \end{subequations}
    Minimizing the residual ($\mathbf{R}$) of $\tilde{\mathbf{K}}_{\Delta}$ over multiple timesteps enables the approximate Koopman operator to identify low-energy modes. These modes, while insignificant for short-term dynamics, may amplify over time and significantly influence long-term behaviour \cite{otto2019linearly}. Further, if we take the case of $\eta =1$, we can observe that minimising equation~\eqref{eq: Koopman Evolution Error} is a linear regression. This regression provides an orthogonal linear projection of $\mathbf{G}_{n+1}^{T}$ onto the range $\mathcal{R}(\mathbf{G}_{n}^{T})$.
    If $\mathbf{G}_{n+1}^{T}$ lies in the linear span of $\mathcal{R}(\mathbf{G}_{n}^{T})$, the residual $\mathbf{R}$ can be minimized to zero. However, if this is not the case, the residual is orthogonal to $\mathcal{R}(\mathbf{G}_{n}^{T})$ and contains the effects of unresolved observables not discovered by the encoder. These effects of unresolved observables are contained in the memory term and the noise term in equation~\eqref{eq:Discrete_Projected_Mori_Zwanzig}. Their models are discussed in the subsequent sections. 
  
\end{itemize}

The overall objective function is a weighted sum of these errors: 
\begin{equation}
    \label{eq: Total objective function}
    \mathcal{J} = \alpha_{rec}\mathcal{J}_{rec} + \alpha_{R}\mathcal{J}_{R}.
\end{equation}
The learnable parameters corresponding to this model include the encoder ($\bm{\theta}_{\mathcal{E}}$), the decoder ($\bm{\theta}_{\mathcal{D}}$), and the approximate Koopman operator ($\bm{\theta}_{\mathbf{\tilde{K}}}$). These weights are obtained by backpropagating through the following loss function,
\begin{equation}
    \label{eq: backprop_objective_KAE}
    \bm{\theta}^{*} = \arg \min_{\bm{\theta}}\mathcal{J}(\bm{\Phi};\bm{\theta}_{\mathcal{E}}, \bm{\theta}_{\mathcal{D}}, \bm{\theta}_{\mathbf{\tilde{K}}}).
\end{equation}
The parameters are then updated in the optimal direction using the ADAM\cite{kingma2014adam} optimization algorithm. The approximate Koopman operator obtained in this manner requires closure through a memory model, which is explained in the subsequent section.

 \subsection{Memory Models}

This work uses two approaches to model the memory term that closes the approximate Koopman operator. In the first approach, we recursively learn the memory convolution kernels through linear regression-based projection while ensuring the generalised fluctuation-dissipation (GFD) constraint is satisfied \cite{lin2022regression}. In the second approach, we model the convolution kernels in recurrent form using RNN without hard-constraining GFD.

 \subsubsection{Generalized fluctuation-dissipation constrained MZ-AE}

In this approach, we build upon the algorithm introduced by Lin et al. \cite{lin2022regression}, adapting it to operate within the latent space of the autoencoder. This algorithm iteratively extracts the memory kernels $\bm{\Omega}_{\Delta (n)}$ using linear regression-based projection and GFD. To describe the algorithm let us begin by setting $n = 0$ in equation~\eqref{eq:Discrete_Projected_Mori_Zwanzig},
\begin{equation}
    \label{eq: discrete reg n=0}
    \hat{\mathbf{g}}_{1} = \tilde{\mathbf{K}}_{\Delta}\hat{\mathbf{g}}_{0} + \mathbf{F}_{0}.
\end{equation}
We already have the approximate Koopman operator $\tilde{\mathbf{K}}_{\Delta}$ by optimizing~\eqref{eq: backprop_objective_KAE}. Further, from equation~\eqref{eq:simple Koopman} we can obtain the orthogonal dynamics as the residual, $\mathbf{F}_{0} = \mathbf{r}_0$. In a similar manner, we can obtain the orthogonal dynamics at $n^{th}$ time step as,
\begin{equation}
    \label{eq: residual_as_orthogonal_dynamics}
    \mathbf{F}_{n} = \hat{\mathbf{g}}_{n+1} - \left( \tilde{\mathbf{K}}_{\Delta}\mathbf{\hat{g}}_{n}+\sum_{k=1}^{n} \bm{\Omega}_{\Delta(k)} \mathbf{\hat{g}}_{n-k}\right).
\end{equation}
Now, GFD (equation~\eqref{eq: in text GFD}) is used to obtain the memory kernels $\bm{\Omega}_{\Delta(n+1)}$,
\begin{equation}
    \label{eq: GFD_memkernel}
    \bm{\Omega}_{\Delta(n+1)}\hat{\mathbf{g}}_{0} = \mathcal{PK}\mathbf{F}_{n} = \mathcal{P}\left(\mathbf{\hat{g}}_{n+2} - \left( \tilde{\mathbf{K}}_{\Delta}\mathbf{\hat{g}}_{n+1}+\sum_{k=1}^{n} \bm{\Omega}_{\Delta(k)} \mathbf{\hat{g}}_{n-k+1}\right) \right).
\end{equation}

In the above equation, we employ linear regression as the projection operator $\mathcal{P}$, following equations~\eqref{eq:linear regression} and~\eqref{eq: regression projection}. This procedure is systematically applied across the dataset 
$
\mathbf{G}_n$ (equation~\eqref{eq:data matrices}) as follows:
\begin{subequations}
\begin{align}
\mathbf{Y}_{n+1} &= \mathbf{G}_{n+2} - \left( \tilde{\mathbf{K}}_{\Delta}\mathbf{G}_{n+1}+\sum_{k=1}^{n} \bm{\Omega}_{\Delta(k)} \mathbf{G}_{n-k+1}\right), \label{eq:1a} \\
\mathcal{J}_{\Omega_{(n+1)}} &= \frac{1}{b}\|\mathbf{Y}_{n+1}- f\left(\mathbf{G}_{0};\bm{\theta}_{\Omega_{(n+1)}}\right)\|^2_F,\label{eq:1b} \\
\bm{\theta}_{\Omega_{(n+1)}}^{*} &= \arg \min_{\bm{\theta}}\mathcal{J}_{\Omega_{(n+1)}}\left(\mathbf{G};\bm{\theta}_{\Omega_{(n+1)}}\right), \label{eq:1c} \\
\bm{\Omega}_{\Delta(n+1)} &\equiv f\left(.;\bm{\theta}_{\Omega_{(n+1)}}^{*}\right). \label{eq:1d}
\end{align}
\end{subequations}

This process is recursively repeated from $n = 1$ to $n=q$, resulting in a set of 
$q$ memory kernels, denoted $\left\{ f\left(\cdot; \bm{\theta}_{\Omega_{(k)}} \right) \right\}_{k=1}^{q}$ where $q$ represents the memory length. The same family of regression functions $f\left(\cdot; \bm{\theta}\right)$ are used across all iterations which ensures uniformity in the projection operator $\mathcal{P}$.  The modelled latent space dynamics are then inferred using the following expression,
\begin{equation}
    \label{eq: Latent Space Dynamics GFD}
    \hat{\mathbf{g}}^{(pred)}_{n+1} = \tilde{\mathbf{K}}_{\Delta}\hat{\mathbf{g}}_n + \sum_{k=1}^{q} f\left( \mathbf{\hat{g}}_{n-k}; \bm{\theta}_{\Omega_{(k)}}\right).
\end{equation}

By construction, this method enforces orthogonality ($\mathcal{P} \mathbf{F}_n = 0$) and satisfies the GFD relation. Therefore, we refer to this approach for non-linear autoencoders as MZ-AE GFD-constrained or MZ-AE GFDc. 

\subsubsection{Recurrent Neural Network}

In this case, we model the memory-based closure term using an RNN. The key advantage of using an RNN over a time-convolution model is that it requires significantly fewer model parameters by comparison. However, it can not be guaranteed that an RNN-based memory model satisfies GFD. We rely on the universal approximation capability \cite{garzon1999dynamical}
 of RNNs to extract correct memory dependence of the residual of the linear Markov operator. In essence, a linear Markov term ($\tilde{\mathbf{K}}_{\Delta}$) and a linear RNN are employed to model GLE (equation~\eqref{eq:Discrete_Projected_Mori_Zwanzig}). 

The linear RNN estimates the memory term by regressing over the residual of the approximate Koopman operator over time as an estimate of the memory term. The modelled latent space dynamics are then obtained as
\begin{equation}
    \label{eq: Latent Space Dynamics}
    \hat{\mathbf{g}}^{(pred)}_{n+1} = \tilde{\mathbf{K}}_{\Delta}\hat{\mathbf{g}}_n + \bm{\xi}_{n}, 
\end{equation}
where $\bm{\xi}_{n}$ is the output of the memory model which recurs over the past $q$ number of observables to produce
\begin{equation}
    \label{eq:lstm_output}
    \bm{\xi}_{n} = \bm{\Omega}\left(\hat{\mathbf{g}}_{n-1}, ..., \hat{\mathbf{g}}_{n-q};\bm{\theta}_{\bm{\Omega}}\right).
\end{equation}
The memory model learns the residual dynamics by regressing the memory model output, $\bm{\xi}$, with the linear evolution residual ($\mathbf{R}$) obtained in equation~\eqref{eq: Linear Evolution residual data},
    \begin{equation}
        \label{eq: Memory Error}
        \mathcal{J}_{\mathbf{\Omega}} = \frac{1}{\eta b}\sum_{i=0}^{b-1}\sum_{j=1}^{\eta}||\mathbf{R}(:,i,j-1) - \bm{\xi}(:,i,j-1)||^{2}_{2},
    \end{equation}
    where $\bm{\xi}(:,i,j) = \bm{\Omega}\left(\hat{\mathbf{g}}_{i+j-1}, ... , \hat{\mathbf{g}}_{i+j-q}\right)$. We describe the mathematical formulation of RNN in the Appendix \ref{sec:LSTM}. The evolved resolved observables are reconstructed back to the original state space using the non-linear decoder $(\mathcal{D})$. To enforce these dynamics, we learn the autoencoder, the approximate Koopman operator and the RNN simultaneously by optimizing a weighted sum of autoencoder reconstruction error $\mathcal{J}_{rec}$ (equation~\eqref{eq:Reconstruction_Loss}), linear evolution error $\mathcal{J}_{R}$(equation~\eqref{eq: Koopman Evolution Error}) and the RNN error $\mathcal{J}_{\Omega}$, 
\begin{equation}
    \label{eq: Total objective function}
    \mathcal{J} = \alpha_{rec}\mathcal{J}_{rec} + \alpha_{R}\mathcal{J}_{R} + \alpha_{\Omega}\mathcal{J}_{\Omega}.
\end{equation}
where the weights $\alpha_{rec}$, $\alpha_{R}$ and $\alpha_{\Omega}$ are tuned for optimal prediction. The model parameters are learned by backpropagating through the objective loss function $ \mathcal{J}$
\begin{equation}
    \label{eq: Final_backprop_objective}
    \bm{\theta}^{*} = \arg \min_{\bm{\theta}}\mathcal{J}(\bm{\Phi};\bm{\theta}_{\mathcal{E}}, \bm{\theta}_{\mathcal{D}}, \bm{\theta}_{\Omega}, \bm{\theta}_{\mathbf{\tilde{K}}}).
\end{equation}
The parameters are then updated in the optimal direction using the ADAM\cite{kingma2014adam} optimization algorithm. Since this method does not guarantee GFD we call this approach MZinspired-AE with Linear RNN or MZi-AE LRNN.\\

MZ-AE GFDc and MZi-AE LRNN operate linearly within the latent space of a non-linear autoencoder, yielding an approximation of the linear GLE. Therefore, it should be emphasized that these approaches are restricted to systems with a single equilibrium point since a single Koopman operator can not capture dynamics across multiple fixed points \cite{page2019koopman, cenedese2022data}. To generalize this approach, we also propose an extension for systems with multiple equilibrium points, where nonlinearity is necessary to capture transitions between them. We introduce this nonlinearity in the memory term using a nonlinear RNN, specifically a long short-term memory (LSTM) network (see Appendix \ref{sec:LSTM}). This extended model is referred to as MZi-AE non-linear RNN (MZi-AE NRNN). This method does not explicitly model the linear GLE (equation~\eqref{eq:Discrete_Projected_Mori_Zwanzig}), and is an extension of MZi-AE LRNN. For a detailed explanation of the motivation behind this choice, we refer the readers to Appendix \ref{sec:Multiple Eqbm Points}.

Building on the approaches described above, it is important to highlight that all these methods execute multi-step prediction entirely in the latent space of the autoencoder, i.e., in the observable space. For instance, when using only the approximate Koopman operator, we compute \(\mathcal{E} \circ \tilde{\mathbf{K}}_{\Delta}^n \circ \mathcal{D}\), where the time evolution occurs solely within the low-dimensional latent space. Here, the encoder \(\mathcal{E}:\mathcal{M} \rightarrow \mathbb{R}^{r}\) maps the state space \(\mathcal{M} \subseteq \mathbb{R}^N\) to a subset of the infinite Hilbert space of observables \(\mathbb{R}^{r} \subset \mathcal{H}\). The approximate finite Koopman operator then acts on these observables to evolve them \(n\) steps ahead in time within the observable space (latent space)  \( \in \mathbb{R}^{r}\). During this process, the observables are not reconstructed back to the state space \(\mathcal{M}\). Once the prediction in the observable space is complete, the decoder \(\mathcal{D}:\mathbb{R}^{r} \rightarrow \mathcal{M}\) reconstructs the predicted observables from the observable space back to the state space.

We note that another approach proposed by Menier \emph{et al}. \cite{menier2023interpretable} shares similarities with the MZ-AE framework, as both leverage the MZ formalism to close the approximate Koopman operator in the latent space of non-linear autoencoders. However, in their work, an alternative strategy is adopted for modeling the memory term. Further, the linear memory model is developed without enforcing the GFD constraint.

 \subsection{Model Selection}

 The proposed model has several hyper-parameters which need to be carefully chosen for optimal performance. Primarily, a systematic grid search method was used to determine the neural network architecture of the encoder, decoder, and RNN. The first key parameter is the number of resolved observables ($r$). In practice, it is decided empirically by evaluating the prediction error of the model for an increasing number of observables until it converges. In this study, we searched for the number of observables close to the inherent dimensions of the model. The prediction horizon ($\eta$) is the number of timesteps over which the approximate Koopman operator is allowed to predict before the memory model learns the obtained residual. It was also chosen using the grid search method.
 
For MZ-AE GFDc, the key hyper-parameter is the memory length which is represented by the number of memory kernels $q$. Increasing  $q$  enhances prediction accuracy but comes at the cost of increased computational complexity. This is highlighted in the study across varying values of \( q \) and \( r \), where we plot the relative mean square error (RMSE) for MZ-AE GFDc and MZi-AE LRNN applied to the limit cycle of cylinder flow (Figure \ref{fig: qrstudy}). Similarly, for the RNN (both linear RNN and LSTM), the two important hyper-parameters are the memory length ($q$) and the number of hidden units ($N_{hu}$). The number of hidden units is the size of the hidden state in RNN that are responsible for carrying the memory information. The memory length ($q$) of the RNN is defined as the number of past observables that the RNN model recurs over to predict the residual of the approximate Koopman operator.  A grid search is done over different values to converge onto a suitable pair of $q$ and $N_{hu}$. An important thing to note is that different dynamical systems have different memory lengths, therefore memory length of the RNN needs to be carefully chosen. A small value of $q$ could lead to loss of information while a large value could increase the computational complexity leading to difficulty in convergence while training.

\section{Numerical Experiments}

\label{sec: Num_Exp}

The MZ-AE framework presented in \S\ref{sec:Methodology} is demonstrated on the two-dimensional flow over a cylinder. In addition, we evaluate the performance of the model on a chaotic Kuramoto-Sivashinsky system. The model parameters and the computational cost for these test cases are in Appendix \ref{app:hyperparameters}.

We examine MZ-AE with memory modelled using a GFD-constrained method (MZ-AE GFDc), with linear RNN (MZi-AE LRNN) and non-linear LSTM (MZi-AE NRNN). To assess the impact of the memory model, we examine the performance of MZ-AE in comparison to an approximate Koopman operator obtained using dynamic mode decomposition (DMD) in the state space. We use the PyDMD module \cite{ichinaga2024pydmd} for implementing the DMD algorithm. We also compare with the approximate Koopman operator obtained in the latent space of the non-linear autoencoder without the memory model, which we refer to as Koopman-autoencoder (Koopman-AE). \\

\subsection{Cylinder Flow Wake}

\begin{figure}[t]
    \centering
    \begin{subfigure}[b]{0.7\textwidth}
        \centering
        \includegraphics[width=\textwidth]{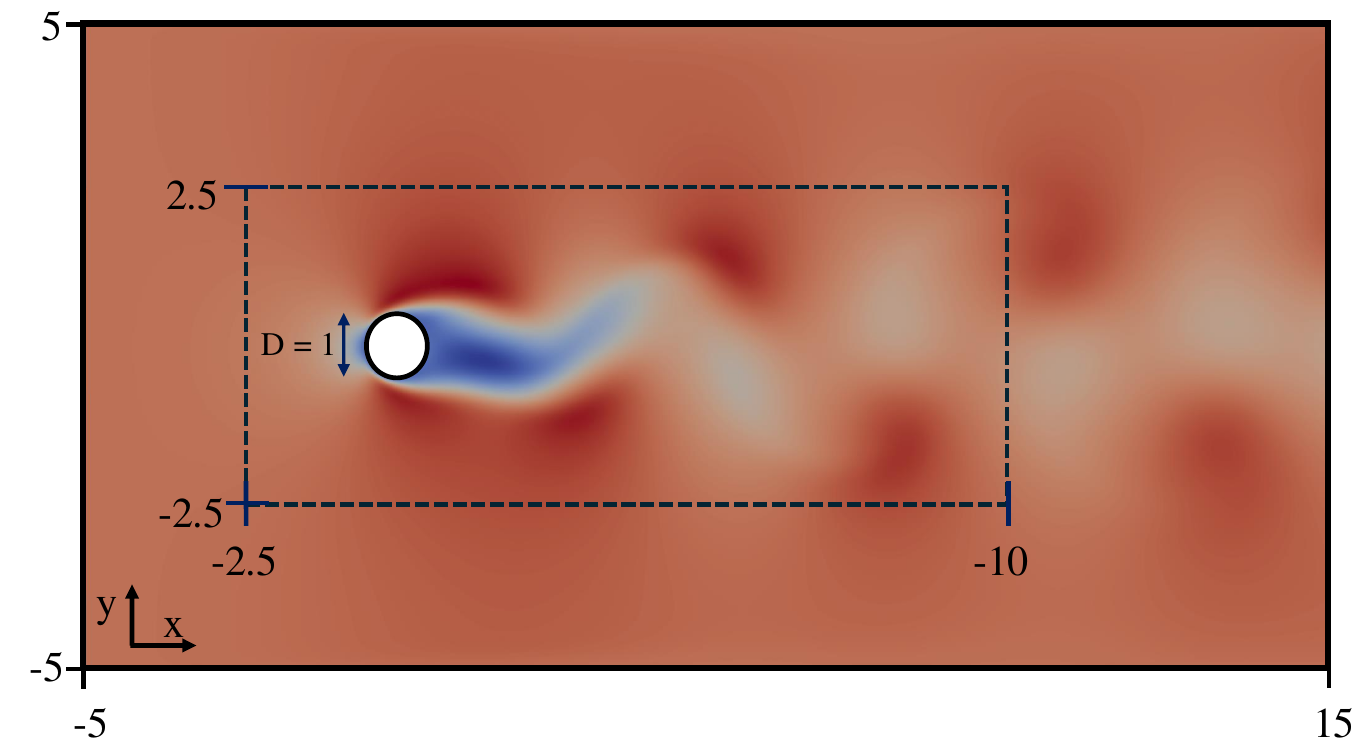}
    \end{subfigure}
    \hfill
    \begin{subfigure}[b]{0.55\textwidth}
        \centering
        \includegraphics[width=\textwidth]{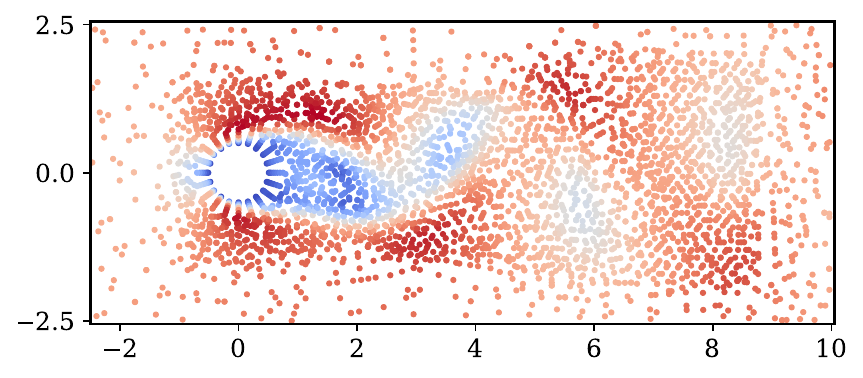}
    \end{subfigure}
    \caption{Top: Computational domain and sampled sub-domain (dotted rectangle) for cylinder flow. Bottom: Data points sampled from the sub-domain for training.}
    \label{fig: Cylinder Domain}
\end{figure}
\begin{figure}[t]
    \centering
    \begin{subfigure}[b]{0.49\textwidth}
        \centering
        \includegraphics[width=\textwidth]{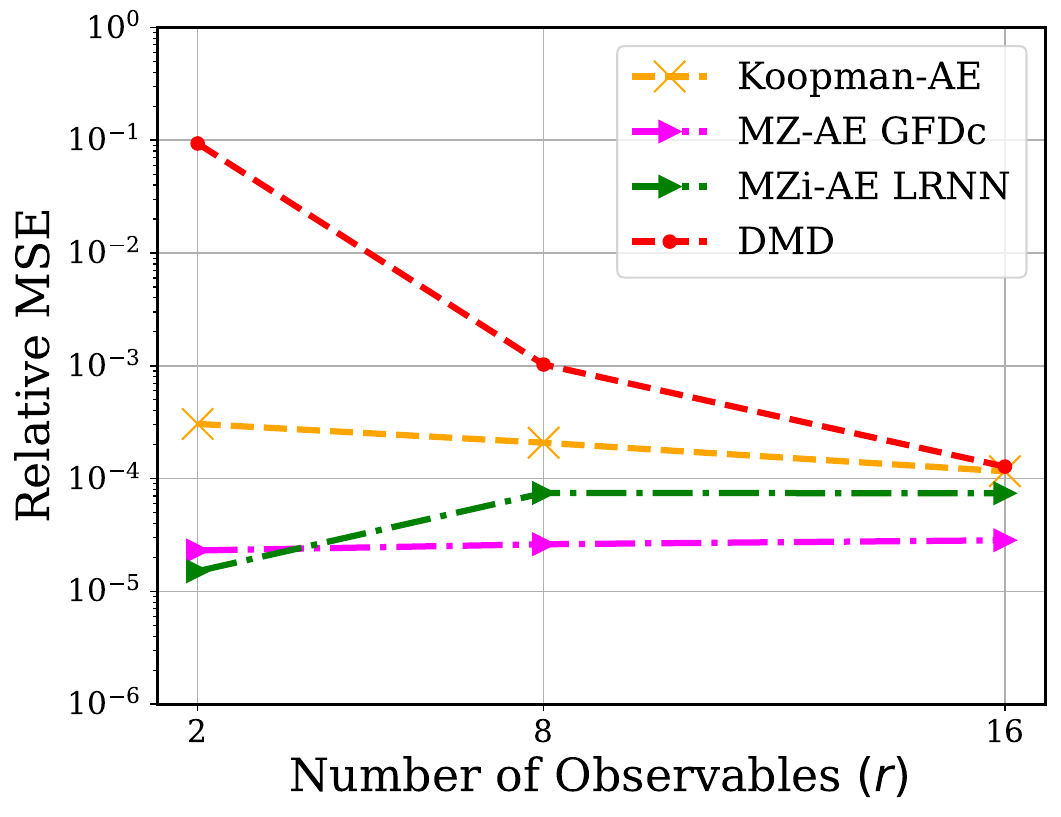}
    \end{subfigure}
    \hfill
    \begin{subfigure}[b]{0.49\textwidth}
        \centering
        \includegraphics[width=\textwidth]{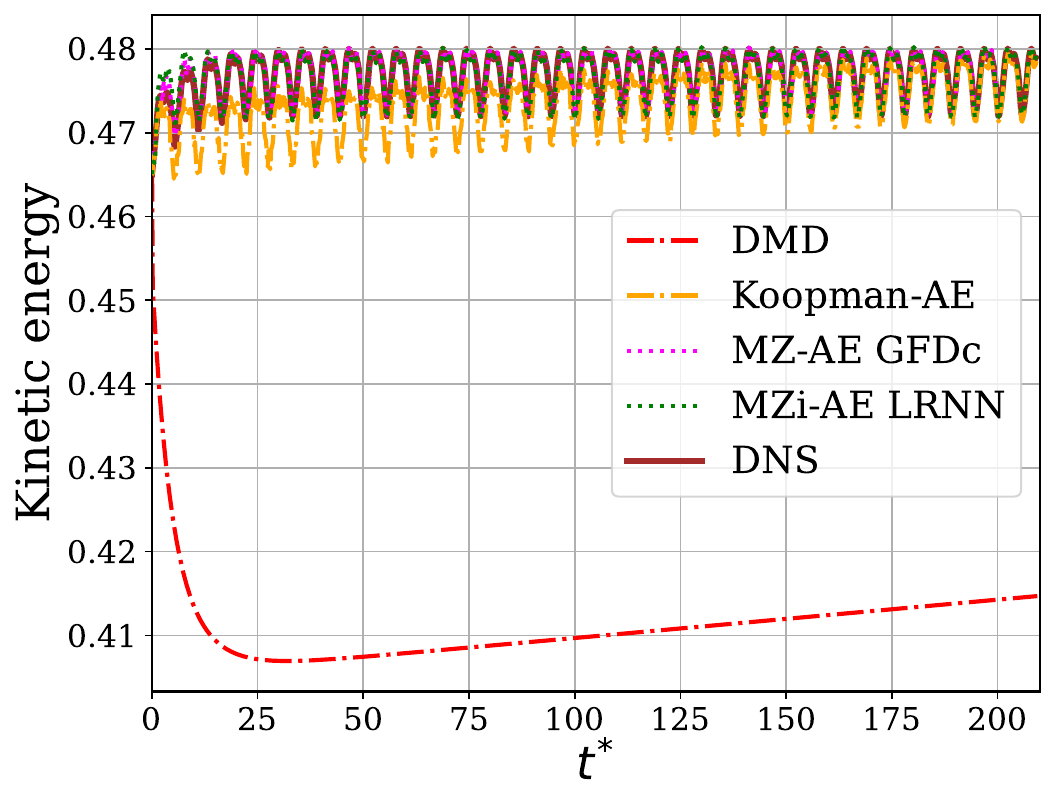}
    \end{subfigure}

    \begin{subfigure}[b]{\textwidth}  
        \centering
        \includegraphics[width=\textwidth]{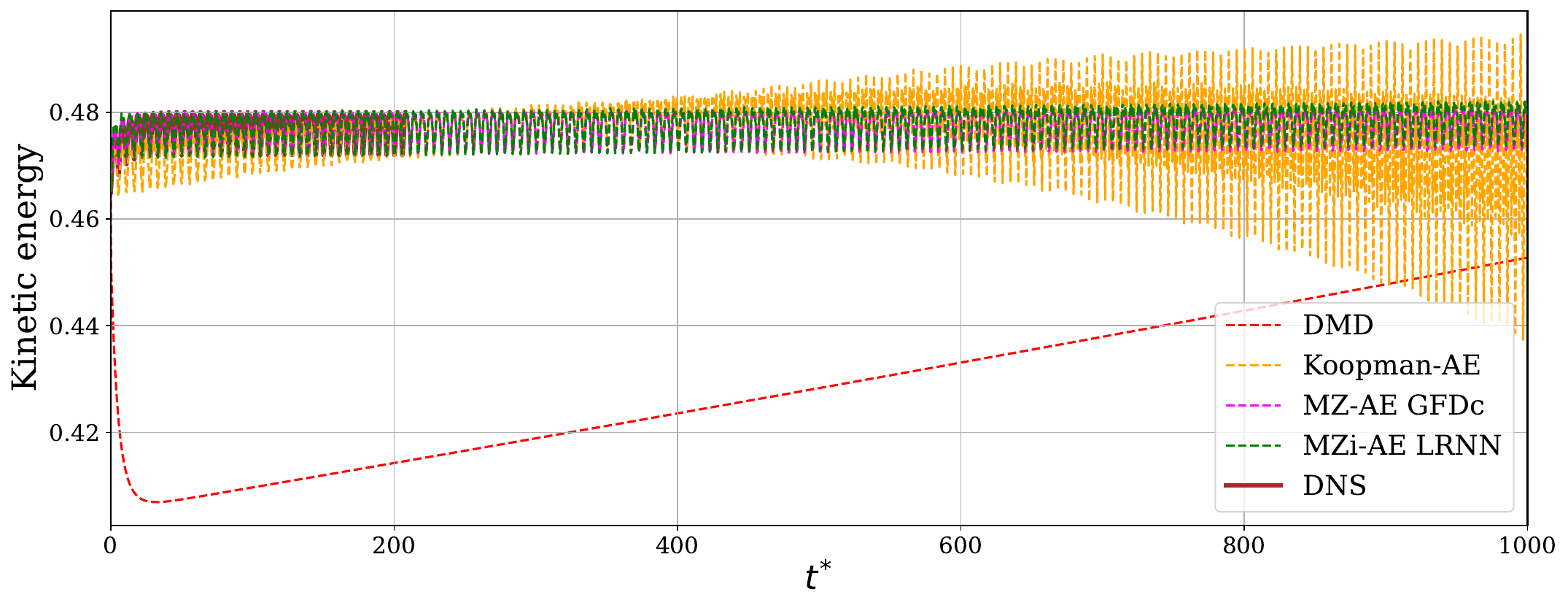} 
    \end{subfigure}
    \caption{Comparative study across different models in the limit cycle regime of the cylinder flow (Case I). Top-left: RMSE of the predicted trajectories with respect to the true solution (direct numerical simulation (DNS)) with increasing number of observables ($r$). Top-right: kinetic energy predictions in state space across different models with $r=2$ resolved observables. Bottom: long-term kinetic energy trajectories comprising 1000 timeunits (corresponding to 180 vortex shedding cycles) of state variables. The x-axis represents normalized time $t^*=n\Delta tD/U_{\infty}$.}
    \label{fig: Limit Cycle Study}
\end{figure}
To probe the workings of the proposed model-reduction
architecture, we first consider the Karman vortex street in the wake
of the cylinder. It is a suitable first test case since it
represents an intrinsically low-dimensional
system~\cite{noack2003hierarchy} expressed in high-dimensional
snapshots. The cylinder flow undergoes a Hopf bifurcation where the trajectory transitions from an unstable equilibrium (baseflow) to the attractor (limit cycle). For this flow, we consider two cases: (i) the trajectory restricted to the attractor (limit cycle), and (ii) the trajectory spanning from the base flow to the limit cycle. It is important to note that the linear GLE should only be valid throughout the trajectory in the first case, as it involves a single equilibrium point. In the second case, two equilibria points are involved,  therefore, we expect a non-linear memory extension to allow for a transition between them. Through our proposed model, we aim
to compress the dynamics to a minimum number of observables
while ensuring satisfactory predictive accuracy over multiple
time steps. 

\begin{figure}[t]
    \centering
    \includegraphics[width=0.9\linewidth]{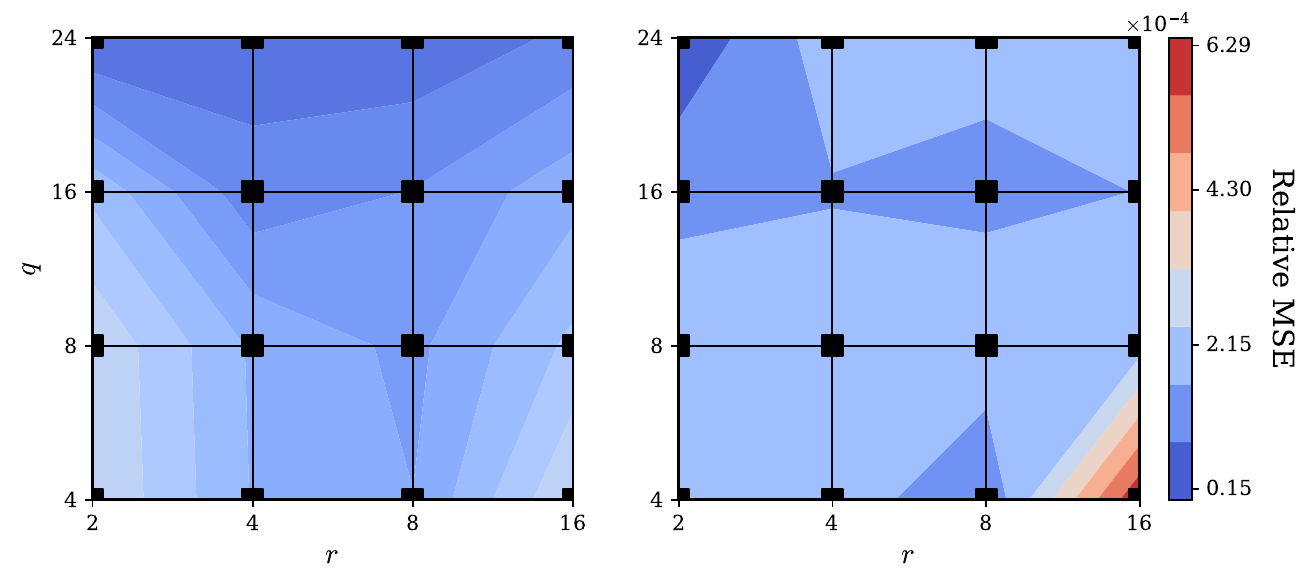}
    \caption{Relative mean squared prediction errors across different numbers of resolved observables ($r$) and the memory length ($q$) in the limit cycle of cylinder flow (Case I). Left: MZ-AE GFDc, Right: MZi-AE LRNN.}
    \label{fig: qrstudy}
\end{figure}

The training data is obtained using the high-fidelity solver
-- {\tt{Nektar++}}~\cite{cantwell2015nektar++} based on a spectral/hp
element framework for Reynolds number $Re = 100$. The baseflow for the simulations was obtained using the selective frequency damping (SFD) method~\cite{jordi2015adaptive} in {\tt{Nektar++}}. The flow was simulated for a cylinder of unit diameter ($D$) in a domain of length $-5D$ to $15D$ in the streamwise direction and $-5D$ to $5D$ in the
normal direction. The velocity snapshots were created by uniformly
sampling the streamwise and normal components of the velocity from a
smaller sub-domain with a streamwise and normal length of $-2.5D$ to
$10D$ and $-2.5D$ to $2.5D,$ respectively. We chose this smaller subdomain since the key dynamics are primarily present in the \textit{wavemaker region}. This region has active physical mechanisms that drive the self-sustained flow oscillations downstream. It consists of the bubble region in the wake of the cylinder which is sufficient for capturing the underlying dynamics of the flow \cite{marquet2008sensitivity}. Figure \ref{fig: Cylinder Domain} visualises the simulation domain, the smaller sub-domain and the sub-sampled points from the domain. The velocity components
were then concatenated into a state vector with $7738$ components. For case-I, we consider a total of $2000$ such velocity snapshots at time
intervals of $0.1D/U_{\infty}$, where $U_{\infty}$ is the free
stream velocity of the flow. For training, $1000$ snapshots were sampled
at a time interval of $0.2D/U_{\infty}$, and the remaining snapshots
were used for testing the model. For case-II, $2800$ velocity snapshots were taken and were split into test and training data similar to the limit-cycle case such that there are $1400$ train and test snapshots. This allowed us to test the
model's prediction capability over the complete trajectory from the unstable equilibrium to the limit cycle.

\subsubsection{Case I: limit cycle}

We begin by analyzing the case where the trajectory is confined to the limit cycle. The prediction horizon of $\eta = 10$ timesteps is taken for all the cases. It can be observed from Figure \ref{fig: qrstudy} that increasing the memory length $q$ improves the relative mean squared error (RMSE) of the prediction for both MZi-AE LRNN and MZ-AE GFDc. We also noted that LRNN struggled to retain information over longer memory lengths due to the vanishing gradient problem.

Figure \ref{fig: Limit Cycle Study} [Top-left] illustrates RMSE of the predicted trajectories within the limit cycle, comparing different models as the number of observables $(r)$ increases, for $r \in [2,8,16]$. For DMD, we use POD modes as resolved observables. We observe that just two POD modes are insufficient for DMD to capture the limit cycle dynamics. As anticipated, a significant reduction in RMSE can be seen with an increasing number of POD modes for DMD. Notably, Koopman-AE demonstrates a three-order-of-magnitude improvement in RMSE compared with DMD with just 2 resolved observables, indicating that the non-linear autoencoder effectively identifies critical observables associated with the underlying dynamics. Furthermore, both MZ-AE GFDc and MZi-AE LRNN achieve an additional order-of-magnitude improvement in RMSE, underscoring the importance of the memory model in providing the necessary approximate closure to enhance predictive accuracy. Further, Figure~\ref{fig: Limit Cycle Study} [Top-left] suggests that the improvement attributed to the memory term is most pronounced when the number of resolved observables is small. As the number of observables increases to 16, the RMSE values across the models converge to nearly identical levels. This convergence is likely a result of the decreased influence of unresolved observables, as increasing the number of resolved observables effectively diminishes the memory model's impact. 

We visualize kinetic energy of the short-term trajectories predicted over 200 time units (approx. 36 vortex shedding cycles) with $r=2$ resolved observables in Figure \ref{fig: Limit Cycle Study} [Top-right] and long-term trajectories over 1000 time units (approx. 180 vortex shedding cycles) in Figure \ref{fig: Limit Cycle Study} [Bottom]. Once again, we observe that a rank-2 DMD fails to accurately capture the system's dynamics, whereas Koopman-AE, MZ-AE GFDc, and MZi-AE LRNN demonstrate strong predictive capabilities. This emphasizes the importance of a non-linear auto-encoder for aggressive model order reduction. However, if we look at the complete trajectory we observe that Koopman-AE is unable to capture the initial asymptotic relaxation of the trajectory to the limit cycle with just 2 resolved observables. Moreover, the long-term behaviour is unstable as illustrated in Figure \ref{fig: Limit Cycle Study} [Bottom]. On the other hand, with the help of the memory correction term, both MZ-AE GFDc and MZi-AE LRNN can capture the complete trajectory and are stable on the limit cycle over a longer number of time steps.

Finally, we observe the eigen-spectrum of the MZ-AE Markov operator. The continuous time eigenvalues ($\lambda$) are obtained from discrete-time eigenvalues ($\mu$) using the relation $\lambda = log(\mu)/\Delta t$. We find that $Img(\lambda)$ of Koopman-AE, MZ-AE GFDc and MZi-AE LRNN models are consistent with the dominant vortex shedding frequency of the limit cycle at $Re = 100$ where the Strouhal number $St = \lambda D / 2 \pi U_{\infty} = 0.179$. The $Real(\lambda)$ for both the linear MZ-AE models comes out to be $8e-5$ indicating that the growth rate is negligible and the eigenvalues are marginally stable.

\subsubsection{Case II: base flow to limit cycle}

\begin{figure}[t]
    \centering
    \includegraphics[width=\textwidth]{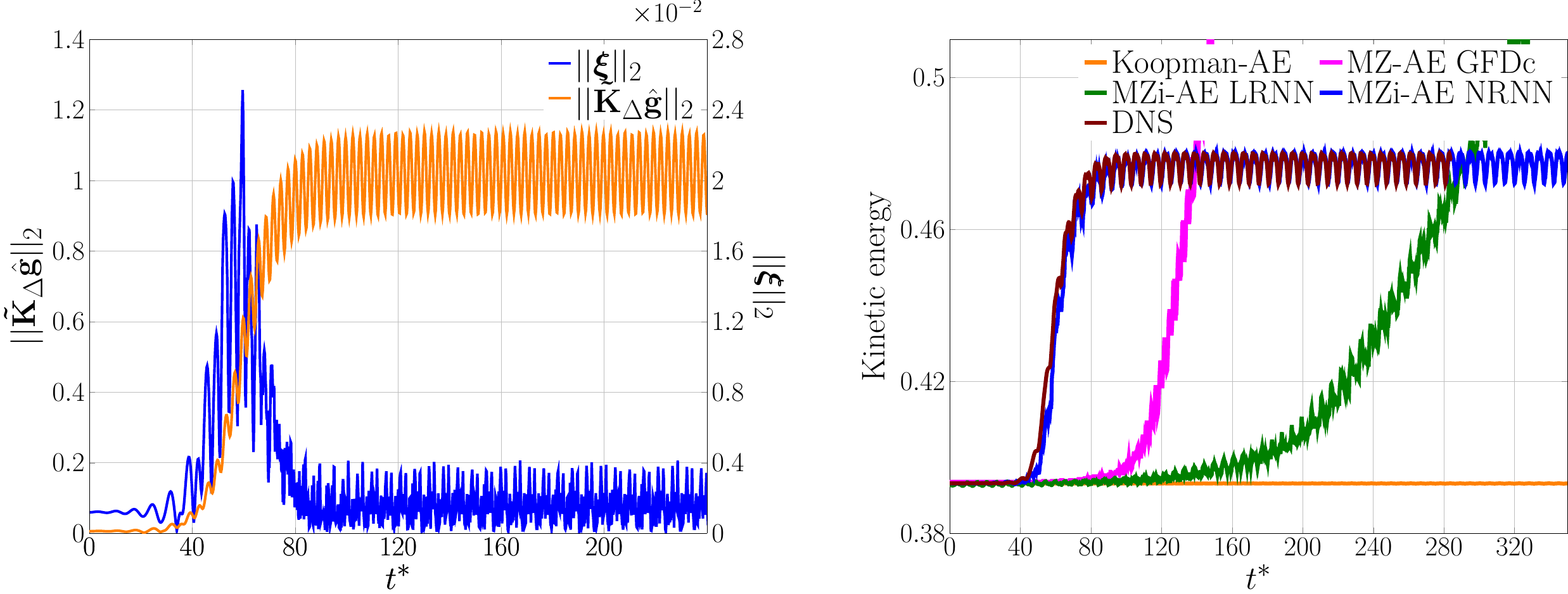}
    \caption{Comparative study across different models from the baseflow to the limit cycle (Case II) using $r=2$ resolved observables. Left: For MZi-AE NRNN,  $L^2$ norm of the contributions to the observable dynamics from the MZ-AE Markov operator ($\mathbf{\tilde{K}}_{\Delta}\hat{\mathbf{g}}$) and the memory model ($\bm{\xi}$). Right: kinetic energy of predicted trajectories in the state space across different models and the true solution (DNS). The x-axis represents normalized time $t^*=n\Delta tD/U_{\infty}$.}
    \label{fig: Trajectory Comparison}
\end{figure}

In this case, we train the model on the complete trajectory from near the unstable baseflow to the attractor which is the limit cycle (see Figure \ref{fig: Trajectory Comparison} [Right] direct numerical simulation (DNS) timeseries). We used the nonlinear autoencoder to extract only two modes (i.e. $r=2$) from the velocity snapshots and expect the memory term to provide necessary corrections for the model to follow the true trajectory. 

From Figure \ref{fig: Trajectory Comparison} [Right], it is evident that Koopman-AE fails to model the transition, as anticipated since two modes are insufficient to capture all the non-linearities present in the system. The linear memory terms in MZ-AE GFDc and MZi-AE RNN attempt to capture transition; however, they fail to converge to the limit cycle. On the other hand, MZi-AE NRNN can forecast the dynamics through the transition regime till the attracting limit cycle and way beyond the training data horizon. This suggests that the non-linear memory model of MZi-AE NRNN provides the necessary non-linearity to capture the transition between unstable baseflow and the limit cycle.

 The observation that the memory correction term facilitates the transition in MZi-AE NRNN is further substantiated by Figure \ref{fig: Trajectory Comparison} [Left]. This figure illustrates the $L^2$ norm of the contributions from both the MZi-AE NRNN Markov operator and the memory model to the observable dynamics. It demonstrates that the memory model is primarily active during the transition regime, while it makes only minor corrections during the limit cycle. Additionally, we observe that the magnitude of the \(L^2\) norm of the contribution from the MZi-AE NRNN Markov operator is two orders of magnitude greater than that from the memory model in the limit-cycle region. These observations suggest that the Markov operator largely contributes to the limit-cycle dynamics while the memory model enables the transition from the unstable equilibrium. 

\subsection{Kuramoto-Sivashinsky}

\label{sec: Full state Kuramoto-Sivashinsky}

The second case we study is the weakly chaotic KS
system which has been developed as a model for flame front
instabilities~\cite{ashinsky1988nonlinear} and has also been used to
model weak fluid turbulence~\cite{hyman1986order}. It is a
popular test case for reduced-order models since obtaining DNS data is
relatively inexpensive, while the equation exhibits a rich and diverse
dynamics at relatively low values of the bifurcation parameter. We
consider the one-dimensional form of the system where the length of
the domain acts as the bifurcation parameter. The KS-system is
governed by a fourth-order partial differential equation, 
\begin{equation}
  \label{eq:KS}
  u_{t} + \lambda u_{xx} + u_{xxxx} + \frac{1}{2}(u_{x})^2 = 0, \quad 0\le x \le L,
\end{equation}
with periodic boundary conditions over the domain length $L$,
\begin{equation}
  \label{eq:KS_BC}
  u(0,t) = u(L,t), u_x(0,t) = u_x(L,t),
\end{equation}
and a user-specified initial condition,
\begin{equation}
  \label{eq:KS_IC}
  u(x,t=0) = u_0.
\end{equation}
The subscripts indicate temporal or spatial derivatives. In
equation~\eqref{eq:KS}, $u_{xx}$ and $u_{xxxx}$ correspond to a
one-dimensional production and dissipation of energy, while $u_x^2$
introduces nonlinearity to the system. We choose a domain length of
$L=22,$ which induces chaotic behaviour with a leading Lyapunov
exponent of $\lambda_1 = 0.043$ and a Kaplan-Yorke dimension $D_{KY}$
of $5.198$~\cite{edson2019lyapunov}. The domain was spatially
discretised using $N=256$ Fourier modes, and the resulting system was
integrated in time using a fourth-order Runge-Kutta implicit-explicit
time stepper~\cite{pachev2022concurrent} with a timestep of $\Delta t
=0.025$. We sample the generated solutions at $\Delta t = 0.25$ and eliminate initial $5000$ snapshots to exclude any
transient data. A total of $8 \times 10^4$ snapshots were created which were divided into training, validation, and testing datasets of size $6 \times 10^4$, $5 \times 10^3$, and $1.5 \times 10^4,$ respectively.

\begin{figure}[t]
    \centering
      \includegraphics[width=0.9\linewidth]{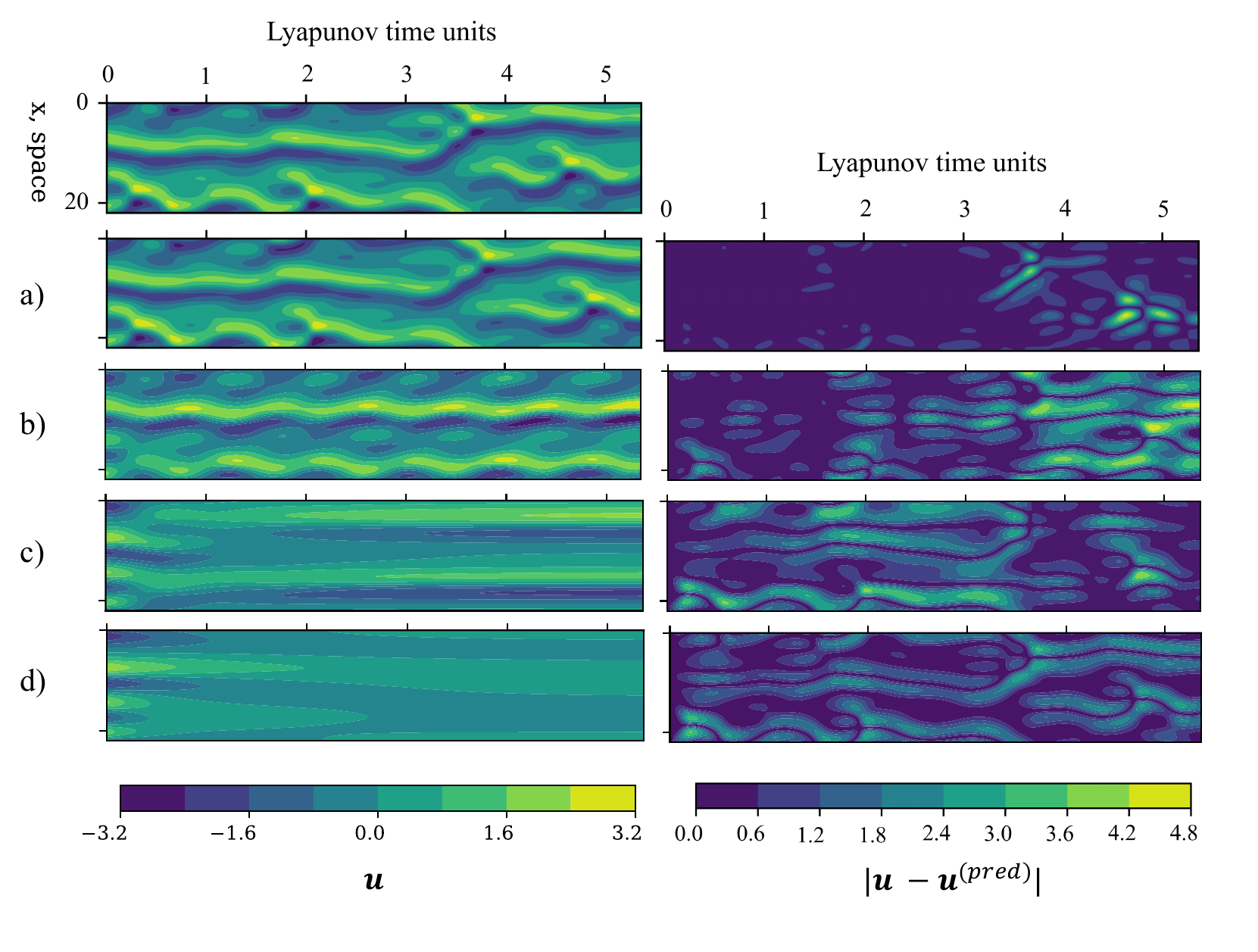}
    \caption{Comparison of trajectories across the models for KS. Top: True solution (DNS). Left: predicted trajectories in the state space, a) MZi-AE NRNN, b) MZ-AE GFDc ($q = 162$), c) MZi-AE LRNN, d) Koopman-AE. Right: absolute error with respect to DNS.}
    \label{fig: KS_Contour_Prediction}
\end{figure}
\begin{figure}[t]
    \centering
    \includegraphics[width=0.9\linewidth]{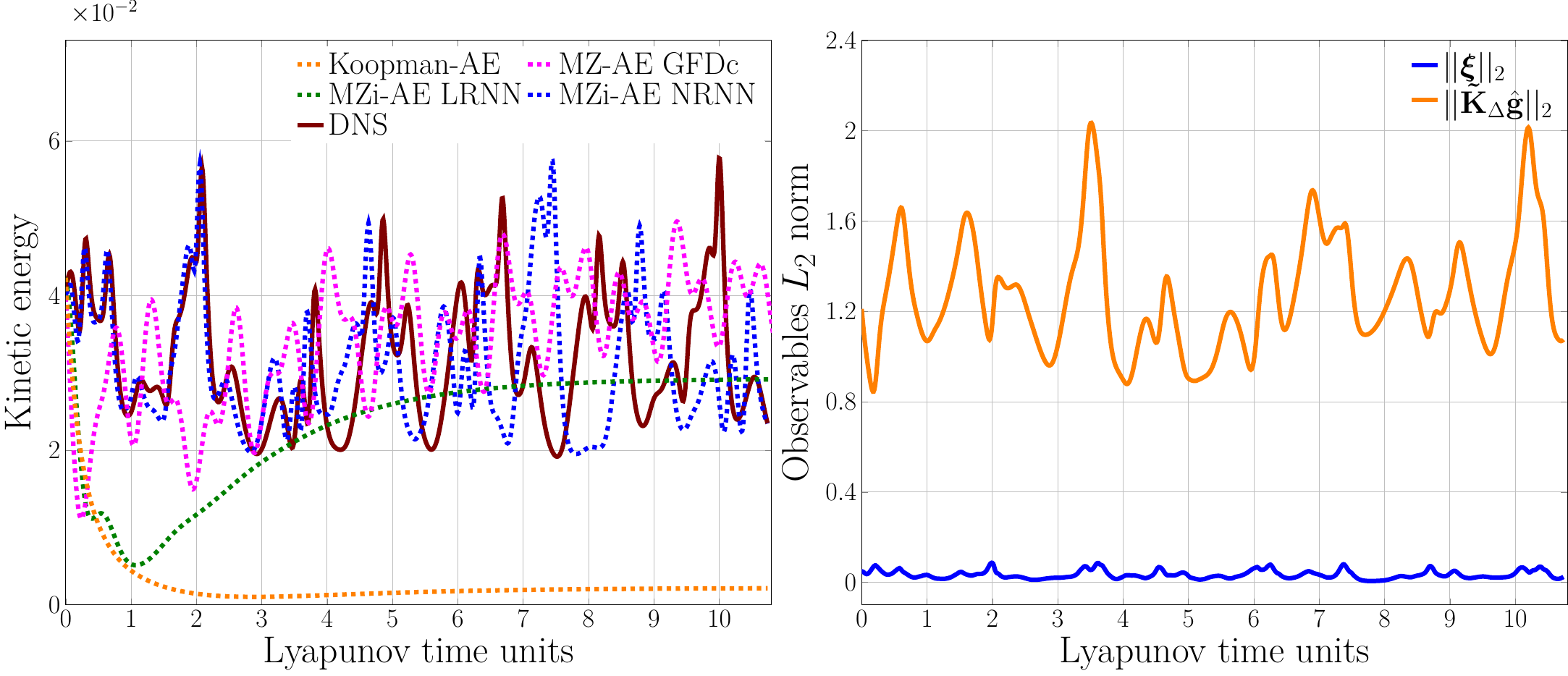}
    \caption{Left: comparison of predicted kinetic energy across models for KS at L = 22 with eight observables.  Right: for MZi-AE NRNN,  $L^2$ norm of the contributions to the observable dynamics from the Markov operator ($\mathbf{\tilde{K}}_{\Delta}\hat{\mathbf{g}}$) and the memory model ($\bm{\xi}$).}
    \label{fig:KS_latent_and_KE}
\end{figure}

For chaotic systems, the Kaplan-Yorke dimension ($D_{KY}$) gives a good estimate of the dimension of the system's attractor. We searched for the number of resolved observables near $D_{KY} = 5.198$ and found 8 observables to be optimal. The prediction horizon ($\eta$) of 20 timesteps and an optimal window length $q$ of $16$ timesteps (corresponding to $\approx 0.13$ Lyapunov Time Units (LTUs)) was taken for both MZi-AE LRNN and NRNN. We present the prediction of the model over $5.5$ LTUs of unseen test data in Figure \ref{fig: KS_Contour_Prediction}. We observe that Koopman-AE struggles to accurately capture the system dynamics, with its predictions quickly damping down close to zero. By contrast, the MZi-AE LRNN demonstrates improved performance for short-term prediction, before eventually converging close to the mean instead of decaying down to zero as can be observed from Figure \ref{fig:KS_latent_and_KE} [Left]. The MZ-AE GFDc model with $q=4$ demonstrates similar performance, with the dynamics converging near the mean (not shown in the figure). In contrast, the MZ-AE GFDc model with $q=162$ better captures some spatio-temporal scales of the system as can be seen in Figure \ref{fig: KS_Contour_Prediction}(b). Although, it still struggles to accurately represent the instantaneous dynamics. Unlike MZ-AE GFDc which is temporal convolution based model, linear RNN was unable to retain information for $q=162$ due to the problem of vanishing gradients while training with RNN. Notably, the MZi-AE NRNN performs significantly better, effectively capturing the dominant characteristics and spatial scales of the system for more than $3$ LTUs. Specifically, it is capable of predicting wave interactions, as evidenced by the contour plots. Nevertheless, as the flow evolves beyond 4 LTUs, the chaotic nature of the system leads to higher absolute errors, a natural outcome given the complexity of the underlying dynamics. The $L^2$ norm of the contribution to the observables for MZi-AE NRNN is presented in Figure \ref{fig:KS_latent_and_KE} [Right] which suggests that the NRNN Markov operator captures the dominant energy in the latent space while the memory model provides low energy corrections to keep it on the desired trajectory.

\begin{figure}[t]
    \centering
    \includegraphics[width=0.9\linewidth]{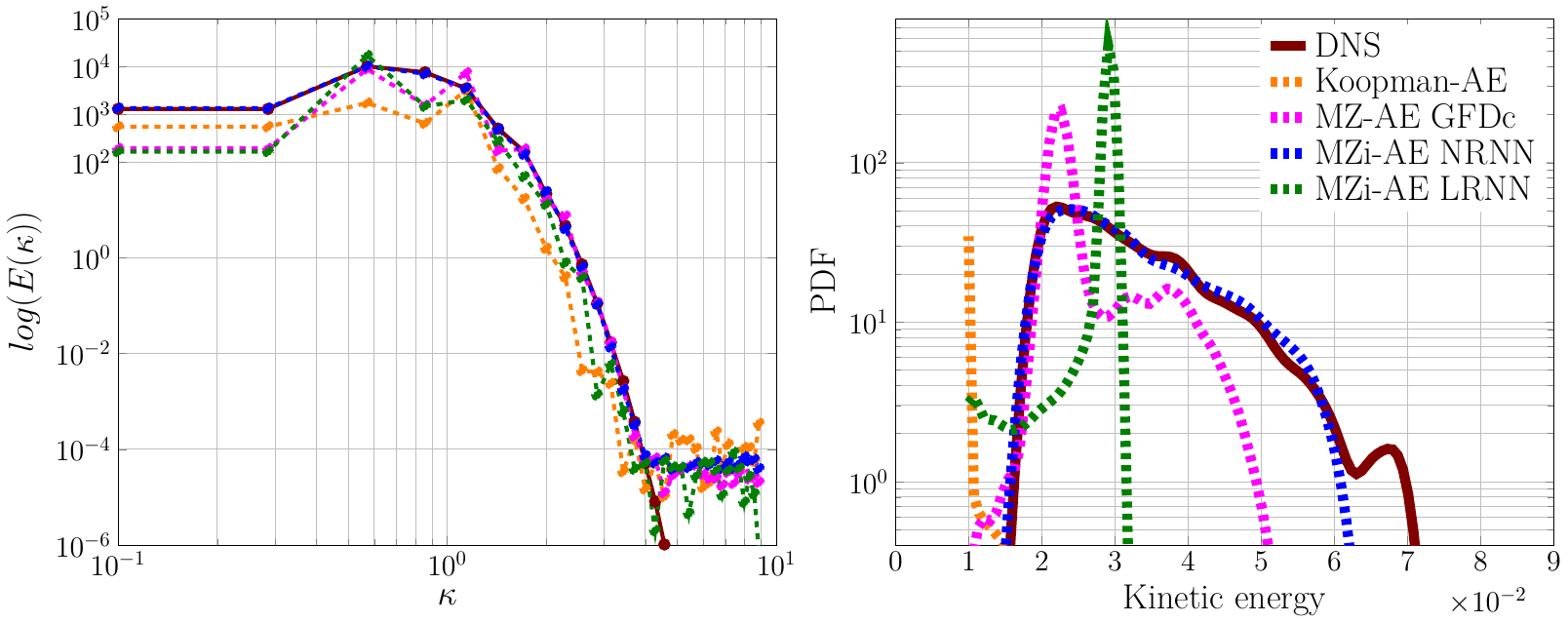}
    \caption{Comparison of long-term prediction statistics for Kuramoto-Sivashinsky system across the models. Left: Kinetic energy spectrum, Right: Kinetic energy probability density function.}
    \label{fig: KS_Stats}
\end{figure}

To investigate the long-term behaviour of the model, we examine the statistics of the system over $160$ LTUs ($15,000$ timesteps) of unseen test data until the statistics converged and are shown in Figure~\ref{fig: KS_Stats}. The time-averaged kinetic energy spectrum of the predictions is presented in Figure~\ref{fig: KS_Stats} [Left] which provides the distribution of energy in different scales of the flow. We can observe that the MZi-AE NRNN is able to accurately capture energy on large scales while MZ-AE GFDc and MZi-AE LRNN deviate less than Koopman-AE from the true spectrum. Further, the MZi-AE NRNN model can capture the long-term statistics of the flow as can be seen from the kinetic energy distribution in Figure~\ref{fig: KS_Stats} [Right]. MZ-AE GFDc and MZi-AE LRNN both converge near the mean of the dynamics over the long term. By contrast, the Koopman-AE model fails to accurately capture the kinetic energy distribution.

\begin{figure}[t]
    \centering
    \includegraphics[width=0.5\linewidth]{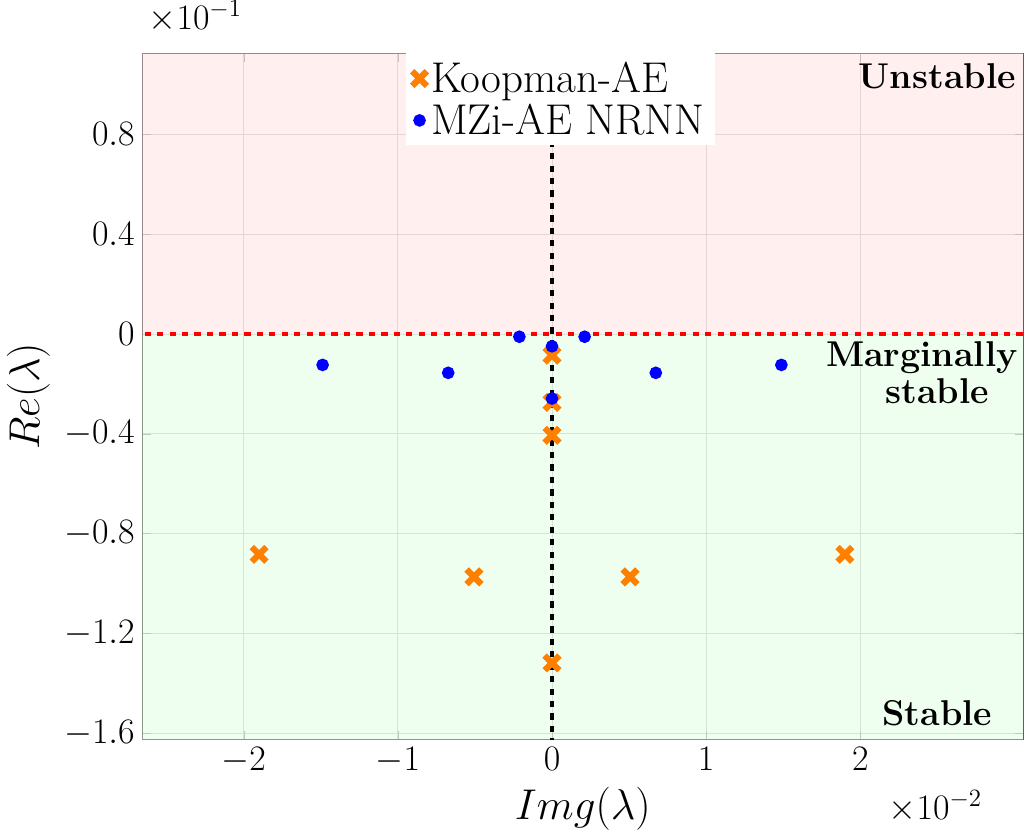}
    \caption{Eigenvalues of the Markov operator for KS system.}
    \label{fig:KS_Eigenvalues}
\end{figure}

 Finally, we analyse the eigenvalues of the MZi-AE NRNN Markov operator and the approximate Koopman operator obtained using Koopman-AE in Figure \ref{fig:KS_Eigenvalues}. It is observed that the majority of the eigenvalues of the MZi-AE NRNN Markov operator have a negative growth rate and are marginally stable which is desired for the dynamics of the attractor of the system. However, the eigenvalues of the Koopman-AE operator are highly damped (larger negative $Real(\lambda)$) which explains the reason behind the decay of the dynamics close to zero. 
 
\section{Conclusion}
\label{sec: Conclusion}
This work proposes a theoretical and computational framework which exploits the
Mori-Zwanzig formalism to furnish a closed-form approximation of the Koopman operator for a nonlinear autoencoder. The framework provides an interpretable ROM while ensuring accurate reconstruction and evolution of the dynamics. We use a non-linear autoencoder to lift the high-dimensional non-linear phase-space dynamics to a low-dimensional approximate linearly invariant latent space. In this latent space, we split the dynamics such that we obtain an approximate Koopman operator for dominant energy observables and introduce a memory term to account for the low-energy residual dynamics. The memory term is modelled using a regression-based projection method which satisfies the GFD constraint in the Mori-Zwanzig formalism \cite{lin2022regression} (MZ-AE GFDc). We also use a linear RNN (MZi-AE LRNN) to model the memory term however, the GFD constraint is not guaranteed to be satisfied in this method.
We demonstrate that these methods provide a substantial improvement over Koopman-AE (Koopman approximation using non-linear autoencoder) and DMD in predicting the limit cycle of cylinder flow using only two resolved observables.

We also propose an extension to MZi-AE LRNN for systems with multiple equilibrium points. For these systems, non-linearity is necessary to capture the transition between the multiple fixed points. We introduce this non-linearity in the memory term by modeling it with a non-linear LSTM. This model is referred to as MZi-AE NRNN. We demonstrate that MZi-AE NRNN is capable of capturing regime transitions from the base flow to the limit cycle in flow past a cylinder, with just two resolved observables. In this case, the linear Markov term captures the limit cycle dynamics, while the nonlinear memory correction term facilitates the transition from the unstable equilibrium. In addition, all the proposed MZ-AE approaches successfully extract the fundamental frequency of the cylinder flow, thereby demonstrating the frameworks' capability to learn the essential dynamical features while maintaining interpretability. 

In another application, we evaluated the model on the multi-scale KS system within the chaotic parameter regime. In the case of a finite and small number of resolved observables, the trajectory produced by Koopman-AE decays down close to the trivial zero solution while that of both MZ-AE GFDc and MZi-AE LRNN settles close to the true mean solution. Both linear memory models struggled with predicting long-term statistics. By contrast, MZi-AE NRNN outperforms these models and shows promising short-term predictability and remarkable long-term statistical performance for considerably low model order. 

We note that the GFD relation which relates the memory term with the noise term can not be guaranteed with RNN \cite{lin2021data_2, lin2022regression} and presents a compelling avenue for future research.  Further investigations could also analyze the performance of MZ-AE when only partial measurements are available in the state-space.

\section*{Acknowledgments}
This work was supported by the Imperial College London - CNRS PhD Joint Program.



\bibliographystyle{unsrt}  
\bibliography{main} 
\appendix

\section{Recurrent Neural Network and LSTM}
\label{sec:LSTM}

\renewcommand{\theequation}{A.\arabic{equation}}

\setcounter{equation}{0}
RNN are a class of neural networks that are recursively applied over a time series to learn the temporal patterns in the dynamical system. An RNN model ($\bm{\Omega}$) with learnable parameters $\bm{\theta}$ takes the input $\mathbf{g} \in {\mathbb{R}^{r}}$ and processes it into an internal or hidden state $\mathbf{h} \in \mathbb{R}^{N_{hu}}$. These hidden states are then recurrently evolved through time from $\mathbf{h}_0$ to $\mathbf{h}_{n}$ using equation  
\begin{equation}
\label{eq: RNN}
        \mathbf{h}_{n} = \mathbf{F}(\mathbf{h}_{n-1},\mathbf{g}_{n-1};\bm{\theta}_{\Omega_1}). 
\end{equation}
The final hidden state $\mathbf{h}_{n}$ is then passed through another dense layer to obtain the desired output $\bm{\xi}_n$
\begin{equation}
    \label{eq: RNN_dense_layer}
    \bm{\xi}_{n} = \mathbf{G}(\mathbf{h}_{n};\bm{\theta}_{\Omega_2}).
\end{equation}
Here, $\bm{\theta}_{\Omega_1}$ and $\bm{\theta}_{\Omega_2}$ are weights of the entire RNN model $\bm{\theta}_{\Omega}$. For a linear RNN, $\mathbf{F}(.)$ and $\mathbf{G}(.)$ are linear functions.
In principle, a RNN is supposed to have infinite memory retention but in practice, they often suffer from long-term memory loss due to vanishing gradients. LSTM (Long short term memory) networks introduced by Hochreiter \emph{et al.} \cite{hochreiter1997long} are a type of gated RNN that are designed to deal with these problems. They use a set of specialised gates and memory cells to store and manipulate information over time. The memory cells employed are; (i) a cell state, $\mathbf{C}_n$, that carries the long-term correlations and (ii) a hidden state, $\mathbf{h}_n$,  that is responsible for short-term trends. Firstly, the input vector $\mathbf{g}$ is mapped to the size of the hidden state using a dense layer. Subsequently, the gates filter information through the following set of equations,
\begin{equation}
\label{eq:LSTM}
\begin{aligned}
    \mathbf{i}_n &= \sigma(\mathbf{W}_i \cdot [\mathbf{h}_{n-1}, \mathbf{g}_{n-1}] + \mathbf{b}_i), \\
    \mathbf{f}_n &= \sigma(\mathbf{W}_f \cdot [\mathbf{h}_{n-1}, \mathbf{g}_{n-1}] + \mathbf{b}_f), \\
    \mathbf{o}_n &= \sigma(\mathbf{W}_o \cdot [\mathbf{h}_{n-1}, \mathbf{g}_{n-1}] + \mathbf{b}_o), \\
    \mathbf{\tilde{C}}_n &= \tanh(\mathbf{W}_c \cdot [\mathbf{h}_{n-1}, \mathbf{g}_{n-1}] + \mathbf{b}_c), \\
    \mathbf{C}_n &=  \mathbf{f}_n \cdot \mathbf{C}_{n-1} + \mathbf{i}_n \cdot \mathbf{\tilde{C}}_n, \\
    \mathbf{h}_n &= \mathbf{o}_n \cdot \tanh(\mathbf{C}_n),
\end{aligned}
\end{equation}

where sigmoid ($\sigma$) and tanh are nonlinear functions. The three gates that are used in an LSTM to control the data flow are: (i) forget gate ($\mathbf{f}_n$) which filters out information to be retained from the cell state, (ii) input gate ($\mathbf{i}_n$) which decides what information should be added to the cell state and, (iii) output gate ($\mathbf{o}_n$) decides the quantity of the current cell state that should be a part of the current hidden state. The set of equations~\eqref{eq:LSTM} are recurred over until the desired time step ($n\Delta t$) is reached which produces the hidden state $\mathbf{h}_{n}$. 

\section{Hyper-parameters for MZ-AE}\label{app:hyperparameters}

\subsection{Cylinder Flow}

The key hyperparameters used for training the models for cylinder flow are listed in tables \ref{tab:table_cyl} and \ref{tab:Cyl_AE}. The computational cost for the models in terms of computational time taken to train the models for cylinder flow with a batch size of $16$ was $\approx 5$ hours over $6,000$ epochs on a RTX6000 GPU. The average computational time, measured over $10$ runs, for predicting $5000$ timesteps (approximately 180 vortex shedding cycles) on the same GPU was $\approx 2.58$ seconds for MZi-AE NRNN, $\approx 2.1$ seconds for MZi-AE LRNN and MZ-AE GFDc. We note that these values are subjective to the hardware availability and quality of algorithm implementation.

\begin{table}[ht] 
    \centering 
    \caption{Training Parameters for Cylinder Wake}
    \label{tab:table_cyl}
    \begin{tabular}{|c|c|c|} 
        \hline
        Hidden and Cell state size & 40  \\
        Batch Size & 16  \\
        Objective Function Weights ($\alpha_{rec}$, $\alpha_{R}$, $\alpha_{\Omega}$) & 10e+2, 1, 1 \\
        Epochs & 6000  \\
        Learning Rate (Adam Optimiser)  & 10e-5\\

        Prediction horizon ($\eta$)  & 10\\
        \hline 
    \end{tabular}
\end{table}
\begin{table}[ht]
    \centering
    \caption{Encoder and Decoder parameters for Cylinder Flow.}
    \label{tab:Cyl_AE}
    \begin{tabular}{|c|c|c|c|}
        \hline
        \multicolumn{2}{|c|}{Encoder} & \multicolumn{2}{|c|}{Decoder} \\
        \hline
        Input Layer & 7738 x 512 & Bottleneck Layer & 8 x 64 \\
        \hline
        Hidden Layer 1 & 512 x 256 & Hidden Layer 1 & 64 x 128 \\
        \hline
        Hidden Layer 2 & 256 x 128 & Hidden Layer 2 & 128 x 256 \\
        \hline
        Hidden Layer 3 & 128 x 64 & Hidden Layer 3 & 256 x 512 \\
        \hline
        Bottleneck Layer & 64 x 8 & Output Layer  & 512 x 7738 \\
        \hline
        \multicolumn{2}{|c|}{Activation Function $\sigma$} & \multicolumn{2}{|c|}{Selu} \\
        \hline
    \end{tabular}
\end{table}

\subsection{Kuramoto-Sivashinsky}

The key hyperparameters used for training the models for cylinder flow are listed in tables \ref{tab:table_KS} and \ref{tab:KS_AE}. The computational cost for the models in terms of computational time taken to train the model for KS with a batch size of $512$ was $\approx 15$ hours over $10,000$ epochs on a RTX6000 GPU. The average computational time, measured over $10$ runs, for predicting $1000$ timesteps (approximately 10.75 LTUs) on the same GPU was $\approx 0.55$ seconds for MZi-AE NRNN, $\approx 0.49$ seconds for MZi-AE LRNN and MZ-AE GFDc.
\begin{table}[h!] 
    \centering 
    \caption{Training Parameters for Kuramoto-Sivashinsky}
    \label{tab:table_KS}
    \begin{tabular}{|c|c|c|} 
        \hline 
        Hidden and Cell state size & 100  \\
        Batch Size & 128  \\
        Objective Function Weights ($\alpha_{rec}$, $\alpha_{R}$, $\alpha_{\Omega}$) & 10e+2, 1, 1 \\
        Epochs & 10000  \\
        Learning Rate (Adam Optimiser) & 10e-5\\

        Prediction horizon ($\eta$)  & 20\\
        \hline 
    \end{tabular}
\end{table}
\begin{table}[h!]
    \centering   \caption{Encoder and Decoder parameters for Kuramoto-Sivashinsky.}
\label{tab:KS_AE}
\begin{tabular}{|c|c|c|c|}
        \hline
        \multicolumn{2}{|c|}{Encoder} & \multicolumn{2}{|c|}{Decoder} \\
        \hline
        Input Layer & 256 x 512 & Bottleneck Layer & 8 x 64 \\
        \hline
        Hidden Layer 1 & 512 x 256 & Hidden Layer 1 & 64 x 128 \\
        \hline
        Hidden Layer 2 & 256 x 128 & Hidden Layer 2 & 128 x 256 \\
        \hline
        Hidden Layer 3 & 128 x 64 & Hidden Layer 3 & 256 x 512 \\
        \hline
        Bottleneck Layer & 64 x 8 & Output Layer  & 512 x 256 \\
        \hline
        \multicolumn{2}{|c|}{Activation Function $\sigma$} & \multicolumn{2}{|c|}{Selu} \\
        \hline
    \end{tabular}
\end{table}

\begin{figure}[t]
    \centering
    \includegraphics[width=0.9\linewidth]{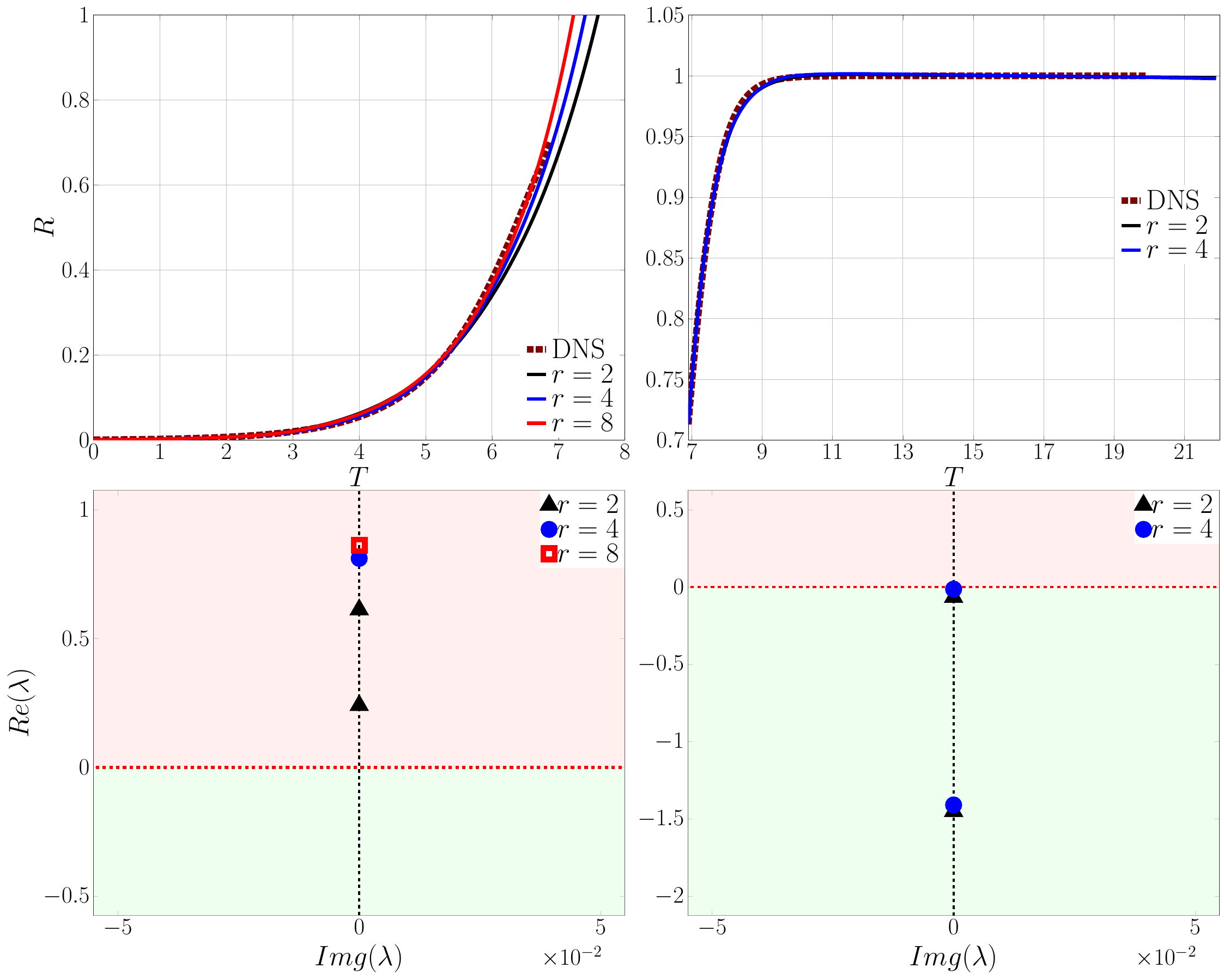}
    
    \caption{Stuart-landau model: Local test in the window of attraction and repulsion without using memory and with increasing number of observables $(r)$. Top-left: Predicted trajectories in the window of repulsion and, Top-right: in the window of attraction. Bottom-left: Unstable eigenvalues at the window of repulsion. Bottom-right: Stable eigenvalues of window of attraction.}
    \label{fig:local_test}
\end{figure}
\newpage
\section{Learning linear representations for systems with multiple fixed points using MZ-AE}
\label{sec:Multiple Eqbm Points}
\renewcommand{\theequation}{C.\arabic{equation}}

\setcounter{equation}{0}

 It has been shown that for non-linear systems with multiple equilibrium points or limit cycles, there is no single Koopman invariant subspace \cite{brunton2016koopman} that globally governs the dynamics. Each equilibrium point is linked to a Koopman expansion, and \textit{cross-over points} exist between these expansions \cite{page2019koopman}. At these cross-over points, the Koopman expansion breaks down, and a new Koopman expansion becomes relevant. As a result, the entire phase space can be partitioned into disjoint invariant regions around equilibrium points, within which the dynamics are topologically conjugate to a linear system \cite{lan2013linearization}.

Dynamical systems with multiple equilibrium points are also termed nonlinearizable systems~\cite{cenedese2022data}. Page and Kerswell \cite{page2019koopman} investigated the Stuart-Landau equation which has one unstable and two stable equilibria. They demonstrated that EDMD is only adept at approximating Koopman expansion within a certain time window of the equilibria that does not contain the crossover point. They found it challenging to obtain a converged linear operator for the trajectories involving the crossover point.  

We recall that linear GLE represents the decomposition of the infinite-dimensional Koopman operator. Consequently, the linear GLE should be valid only in regions where the infinite-dimensional Koopman operator is applicable. Therefore for systems with multiple fixed points, it should be challenging for GLE to capture the dynamics involving transition from one fixed point to another. In this brief investigation, we demonstrate that incorporating a memory term can enhance the linear representation with a finite set of observables. Furthermore, introducing a non-linear memory term provides the necessary non-linearity to accurately capture transitions between fixed points. 

Following the study of Page and Kerswell\cite{page2019koopman}, in this section we study how MZ-AE performs for systems with multiple equilibrium points. We start with the Stuart-Landau equation,
\begin{equation}
    \label{eq:complex_SL}
    \frac{dA}{dt} = a_0A - a_1A|A|^2, \;\;\;\;\; A \in \mathbb{C}, 
\end{equation}
where the complex amplitude can be written in polar coordinates as, $A(t) = |A(t)|exp[i\theta(t)]$. 
We specifically consider the evolution equation for the amplitude $|A(t)|$ normalised by the bifurcation parameter $\mu$,
\begin{equation}
\label{eq:SL}
\frac{d R}{dT} = R - R^3,
\end{equation}
where $R(t) = |A(t)|/\sqrt{\mu}$, ~$T = \mu t$ and $\mu>0$. This equation has three fixed points, two attracting at $R = \pm 1$ and a repelling point at $R=0$. The crossover point between the unstable fixed point and the stable fixed point has been evaluated to be at $R = 1/\sqrt{2}$ \cite{page2019koopman}. The region starting from $R=0$ to the crossover point is referred to as repelling window and the region from cross-over point to $R = 1$ as an attracting window. A total of 1600 snapshots were sampled at $\Delta t = 0.0125$ and were split into train and test data (similar to cylinder case) resulting in a sample rate of  $\Delta t = 0.025$. The repelling window consists of 300 snapshots while the attracting window consists of 500 snapshots.

We begin by analysing the regions of attraction and repulsion surrounding each fixed point individually, using a finite Koopman approximation operator without memory (employing Koopman-AE, as detailed in \S\ref{sec:Methodology}). In Figure \ref{fig:local_test} [Top-left], we observe that an increasing number of observables improves the prediction accuracy in the repelling window. In the attracting window, two observables corresponding to stable eigenvalues are sufficient to capture the dynamics as can be seen from Figure \ref{fig:local_test} [Top-right]. In the repelling window, for greater than two observables ($r > 2$), there is a single unstable eigenvalue (see Figure \ref{fig:local_test} [Bottom-left]) that converges near the unstable eigenvalue associated with the fixed point (\emph{i.e.} $\lambda = $ 1). As expected, the approximated Koopman spectrum converges to the true one as the number of observables increases. The rest of the eigenvalues for $r>2$ are stable and are highly damped (not plotted). Similarly, in the attraction window, the eigenvalue corresponding to $\lambda = 0$ is better approximated by increasing the number of observables from $r=2$ to $r=4$. These observations demonstrate that a finite linear approximation effectively linearizes the repelling window around the unstable fixed point and the attracting window around the stable fixed point. We would like to point out that the Koopman expansion is different about repelling fixed point and the attracting fixed point. 

\begin{figure}[t]
    \centering
    \includegraphics[width=\linewidth]{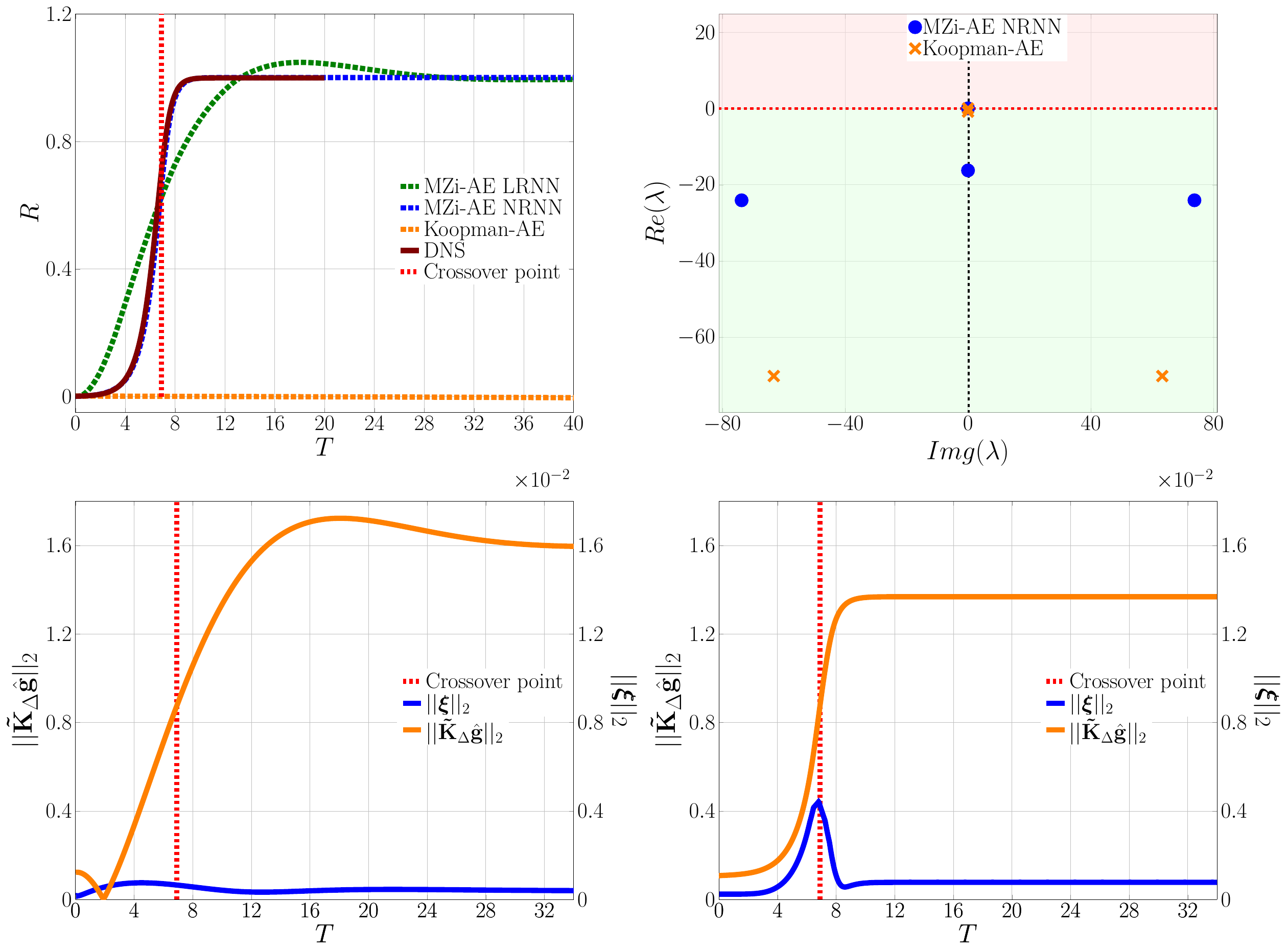}
    \caption{Stuart-landau model: global test with four resolved observables in the complete trajectory from the window of repulsion to the window of attraction using Koopman-AE, MZi-AE LRNN and MZi-AE NRNN. Memory length of 0.4 timeunits is used in the models with memory.  Top-left: Predicted trajectories in the state space. Top-right: Spectrum of the approximate Koopman operator in Koopman-AE, and Markov operator in MZi-AE NRNN. Bottom left: $L^2$ norm of latent space contribution of MZi-AE LRNN Markov operator $(\tilde{\mathbf{K}}_{\Delta}\hat{\mathbf{g}})$ and linear memory model $(\bm{\xi})$. Bottom-right: $L^2$ norm of latent space contribution for MZi-AE NRNN.}
    \label{fig:global_test}
\end{figure}


Next, we examine if the finite Koopman approximation with the limited number of observables can capture the dynamics across the complete trajectory from one fixed point to another. We use Koopman-AE to approximate the Koopman operator with just 4 observables. From Figure \ref{fig:global_test} [Top-left], we are not able to obtain a converged finite Koopman approximation that governs the dynamics across the fixed points with just 4 observables. The finite linear operator obtained by Koopman-AE has two marginally stable and two highly damped eigenvalues (Figure \ref{fig:global_test} [Top-right]) which restricts the dynamics to the fixed point at $R=0$. The prediction accuracy can be improved either by increasing the number of observables or by adding a memory term to account for the unresolved observables. When a linear memory term is added (\emph{i.e.} we use MZi-AE LRNN), we observe improved prediction of the true trajectory. However, the transition regime is still not captured accurately. 

When a non-linear memory correction term is used, we observe convergence of the predicted trajectory to the true trajectory. The linear Markovian term learns the marginally stable eigenvalue associated with the window of attraction while the non-linear memory term provides the required correction by capturing the unresolved dynamics. On top of this, the non-linear memory term provides the non-linearity required for the jump between the window of repulsion and the window of attraction. This is suggested by the fact that the $L^2$ norm of the contribution by the non-linear memory model in the latent space peaks exactly at the crossover point (see Figure \ref{fig:global_test} [Bottom-right]). 

\end{document}